\documentclass[a4paper,fleqn]{cas-dc}

\usepackage{natbib}
\usepackage{subfigure}
\usepackage{hyperref}       
\usepackage{url}            
\usepackage{amsfonts}       
\usepackage{nicefrac}       
\usepackage{microtype}      
\usepackage{xcolor}         
\usepackage{amssymb}
\usepackage{graphicx}
\usepackage{tabularx}
\usepackage{multirow}
\usepackage{paralist}
\usepackage{enumitem}
\usepackage{booktabs}
\usepackage{caption}
\usepackage[switch]{lineno}

\usepackage{bm}
\usepackage[table,dvipsnames]{xcolor}
\usepackage{collcell}
\usepackage{array}
\usepackage{arydshln}
\usepackage{etoolbox}

\usepackage{newunicodechar}
\newunicodechar{−}{-}

\def\tsc#1{\csdef{#1}{\textsc{\lowercase{#1}}\xspace}}
\tsc{WGM}
\tsc{QE}
\tsc{EP}
\tsc{PMS}
\tsc{BEC}
\tsc{DE}

\begin{document} 
\begin{sloppypar}
\let\WriteBookmarks\relax
\def\floatpagepagefraction{1}
\def\textpagefraction{.001}
\shorttitle{SARU: A Shadow-Aware and Removal Unified Framework for Remote Sensing Images with New Benchmarks}
\shortauthors{Zy. Bo, et~al.}

\title [mode = title]{SARU: A Shadow-Aware and Removal Unified Framework for Remote Sensing Images with New Benchmarks}        

\author[1]{Zi-Yang Bo$^\dagger$}[style=chinese]
\ead{e24201058@stu.ahu.edu.cn}

\author[1]{Wei Lu$^\dagger$}[style=chinese]   
\ead{luwei_ahu@qq.com}

\author[2]{Hongruixuan Chen}[style=chinese]
\ead{qschrx@gmail.com}

\author[1]{Si-Bao Chen}[style=chinese]
\cormark[1]
\ead{sbchen@ahu.edu.cn}

\author[1]{Bin Luo}[style=chinese]
\ead{luobin@ahu.edu.cn}

\address[1]{MOE Key Lab of ICSP, Anhui Provincial Key Lab of Multimodal Cognitive Computation, IMIS Lab of Anhui Province, School of Computer Science and Technology, Anhui University, Hefei, China}
\address[2]{Graduate School of Frontier Sciences, The University of Tokyo, Chiba, 277-8561, Japan}

\cortext[cor1]{Corresponding author, $^\dagger$ Equal contribution.}

\begin{abstract}
Shadows are a prevalent problem in remote sensing imagery (RSI), degrading visual quality and severely limiting the performance of downstream tasks like object detection and semantic segmentation. Most prior works treat shadow detection and removal as separate, cascaded tasks, which can lead to cumbersome process and error accumulation. Furthermore, many deep learning methods rely on paired shadow and non-shadow images for training, which are often unavailable in practice.
To address these challenges, we propose Shadow-Aware and Removal Unified (SARU) Framework , a cohesive two-stage framework. First, its dual-branch detection module (DBCSF-Net) fuses multi-color space and semantic features to generate high-fidelity shadow masks, effectively distinguishing shadows from dark objects. Then, leveraging these masks, a novel, training-free physical algorithm (N$^2$SGSR) restores illumination by transferring properties from adjacent non-shadow regions within the single input image.
To facilitate rigorous evaluation and foster future work, we also introduce two new benchmark datasets: the RSI Shadow Detection (RSISD) dataset and the Single-image Shadow Removal Benchmark (SiSRB). Extensive experiments on the AISD and RSISD datasets demonstrate that SARU achieves state-of-the-art shadow detection performance with IoU scores of $86.35\%$ and $89.71\%$, respectively, providing a performance gain of over $3.5\%$ compared to recent methods like CADDN and SDDNet. For shadow removal, our training-free N$^2$SGSR algorithm attains an average processing speed of approximately $1.3$s, which is over $10$ times faster than the state-of-the-art MAOSD while  maintains an SRI value close to 0.9 on both the AISD and SiSRB datasets, a level comparable to the advanced RS-GSSR method. By holistically integrating shadow detection and removal to mitigate error propagation and eliminating the dependency on paired training data, SARU establishes a robust, practical framework for real-world RSI analysis. The source code and datasets are publicly available at: \url{https://github.com/AeroVILab-AHU/SARU}.
\end{abstract}

\begin{keywords}
Shadow detection \\ 
Shadow removal  \\
Remote sensing \\
Dual-branch feature fusion \\
Superpixel segmentation
\end{keywords}

\maketitle

\section{Introduction}
\label{sec:intro}
With the rapid advancement of computer vision and earth observation technologies, remote sensing imagery (RSI) has become an indispensable data source for critical practical applications, including urban planning, environmental monitoring, and disaster damage assessment. However, shadows cast by buildings, trees and other elevated objects present a persistent, unavoidable challenge that severely degrades RSI data quality and undermines the performance of automated information extraction algorithms \citep{10144358}. Shadowed regions trigger substantial losses of true ground reflectance information and severe spectral distortion, which not only weaken the accuracy of target detection and classification tasks, but also easily lead to misjudgments in image interpretation, directly limiting the practical application value of high-resolution RSI \citep{Hufel2024ShadowDA}.
While shadows can be exploited for niche applications such as ground object height estimation \citep{9945884} and 3D urban geometric modeling \citep{5597187}, their detrimental impacts dominate nearly all mainstream RSI downstream tasks. In object detection, for instance, reduced brightness and contrast in shadowed areas heavily hinder the identification of small targets like vehicles and road markings, leading to a sharp rise in missed detection rates \citep{10142024, 11375659, lu2026unravelnet}. In semantic segmentation \citep{DERESSU2025104839,11313649}, road extraction \citep{11021615,yang2026semantic} and change detection \citep{lu2026lwganet, liu2025commonality}, shadows are frequently misclassified as dark objects (e.g., water, dark asphalt) or false change regions caused by varying solar angles. Accordingly, developing robust shadow detection and removal techniques has become a vital preprocessing step to enhance the reliability of downstream RSI tasks, making this research highly meaningful in both theoretical exploration and engineering application.
To tackle the interference of RSI shadows, researchers have developed a series of solutions, with technical routes evolving from traditional physical and handcrafted feature-based methods to deep learning-driven approaches. Early methods were split into two main categories: physical model-based and handcrafted feature-based schemes. Physical model-based methods predicted shadow coverage using solar geometry parameters and Digital Elevation Models (DEMs) \citep{LUO2019197}, or estimated irradiance in shadowed regions via complex illumination models \citep{SILVA2018104}. Though theoretically grounded, these methods rely heavily on high-precision auxiliary data and fail to adapt to the complex geometric layout of urban environments, resulting in limited practical scalability. Feature-based methods, by contrast, extracted distinctive shadow properties in specific color spaces (e.g., HSV, Lab) or texture domains to set thresholds or train classifiers \citep{10.1007/978-3-319-64698-5_26, 11058953}; techniques including histogram matching \citep{10.1007/978-3-319-64698-5_26} and color transfer \citep{10.1080/01431160600954621} were also adopted to correct shadow-induced spectral distortions. These methods boast high computational efficiency, yet they depend excessively on manual feature design, lack robustness under complex backgrounds and fluctuating illumination, and often produce incomplete detection or unnatural restoration artifacts.
In recent years, deep learning has revolutionized shadow processing, with supervised methods based on convolutional neural networks (CNNs) becoming the dominant paradigm in shadow detection and removal \citep{10191081, 10058544, 10205508}. Despite remarkable performance gains, existing deep learning methods still face three critical bottlenecks that restrict their performance on urban RSI:
\begin{itemize}
	\item[1)] \textbf{Error Propagation in Conventional Cascaded Frameworks:} Most existing shadow processing systems follow a standard two-stage cascaded pipeline, treating shadow detection and removal as two separate, sequential tasks \citep{MOVIA2016485, WANG2024110771}. In this structure, the raw output of the detection stage serves directly as the input for the removal stage, with no dedicated error suppression mechanism. Minor flaws in detection, such as blurred shadow boundaries, missed small shadows, or misclassification of dark objects are directly passed to the removal stage and often amplified, severely compromising the final restoration quality.
	\item[2)] \textbf{Domain Gap Between Natural Images and RSI:} Most shadow detection models are developed, trained and validated solely on natural images. Directly transferring such models to RSI completely ignores the unique domain characteristics of RS data \citep{10833852, 10552804}. Compared with natural images, RSI adopt a nadir overhead perspective, possess complex and diverse urban spatial textures, and also contain a large number of ground objects with spectral features highly similar to shadows. The extreme scarcity of large-scale, high-quality annotated datasets tailored specifically for RS scenes is the core bottleneck of shadow detection in RSI, which also leads to unsatisfactory performance of directly transferred models in real-world RS scenarios.
	\item[3)] \textbf{Over-Reliance on Paired Training Data:} Supervised shadow removal models depend entirely on large-scale paired datasets containing shadow and corresponding non-shadow images for training \citep{YUAN2026112001, HUANG2025111126}. However, collecting such perfectly paired data is practically infeasible in real-world RSI scenarios, as it requires capturing the same location at the identical time and illumination condition with and without shadows. Moreover, evaluating the quality of single-image shadow removal in the absence of ground-truth shadow-free references remains another critical challenge. This data limitation drastically weakens the model’s generalization to real RS scenes.
\end{itemize}

\begin{figure*}\centering
	\includegraphics[width=1\linewidth]{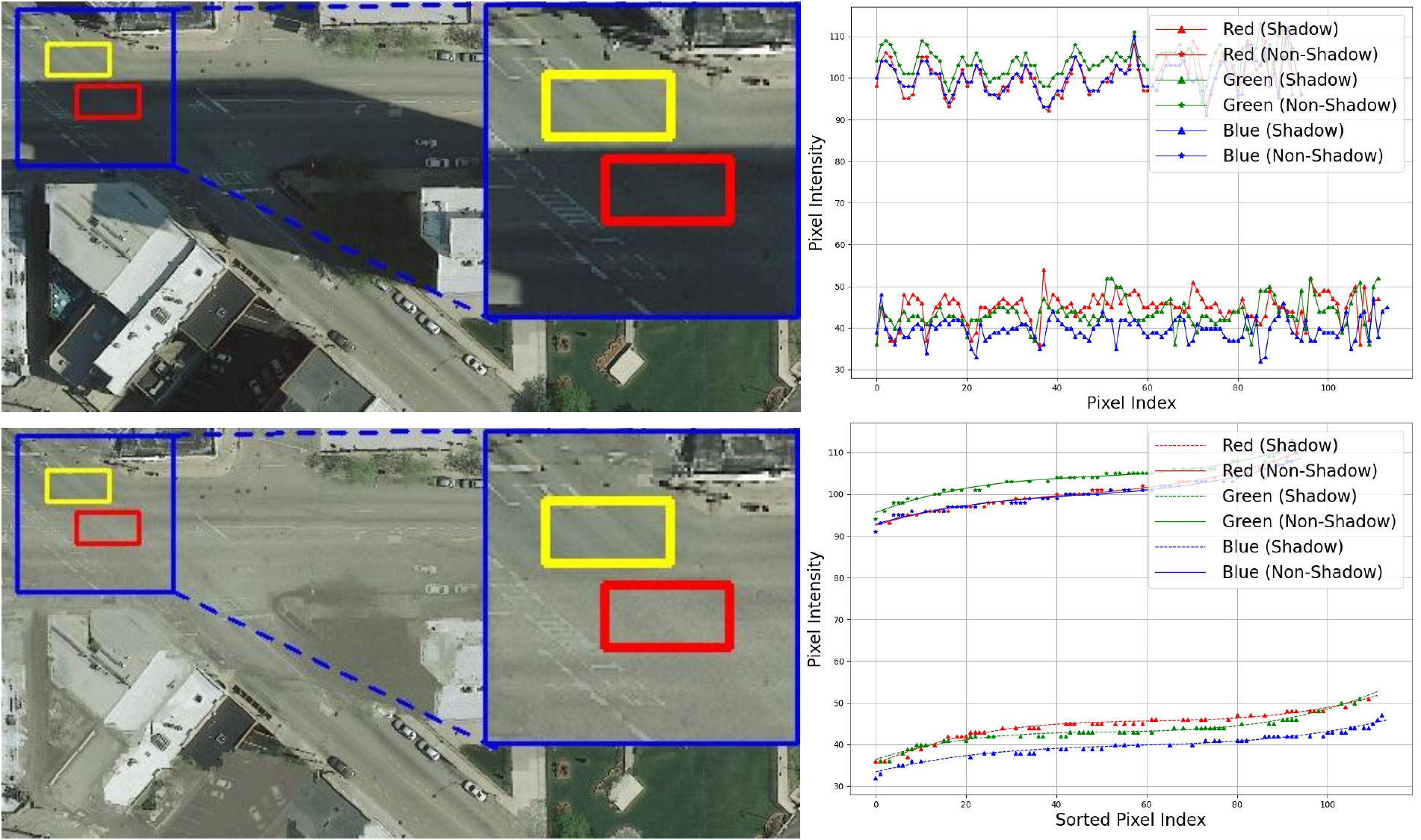}
	\caption{The left image shows a comparison of sampled images before and after shadow removal. The top-right image illustrates that randomly sampled shaded and unshaded regions within the same background exhibit similar pixel intensity distributions across the RGB channels. The bottom-right image demonstrates a similar trend in pixel value variation when sorted in ascending order for both regions. This suggests a potential mapping relationship between shaded and unshaded areas under similar background.}	\label{fig:1}
\end{figure*}

To address these core limitations and adapt to the unique traits of urban RSI, this paper proposes a novel two-stage unified (SARU) framework for joint shadow detection and removal. Distinct from conventional isolated cascaded systems, this framework integrates detection and removal as two closely coordinated, sequential modules with targeted design to suppress cross-stage error propagation, rather than operating as fully decoupled tasks. First, to address the critical shortage of high-quality open datasets in RSI shadow research, we construct and publicly release two new benchmark datasets: the large-scale, diverse urban RSI Shadow Detection Dataset (RSISD) for high-precision shadow detection, and the Single-Image Shadow Removal Benchmark (SiSRB), the first dedicated benchmark for RSI single-image shadow removal evaluation. Building on this data foundation, we design the Dual-Branch Color-Semantic Fusion Network (DBCSF-Net) to generate accurate shadow masks, followed by the Nearest Neighbor Superpixel-Guided Shadow Removal (N$^2$SGSR) algorithm — a training-free, physics-inspired method that eliminates reliance on paired training data. As visualized in Fig.~\ref{fig:1}, N$^2$SGSR restores shadow region illumination using only color and texture cues from adjacent non-shadow superpixels within a single RSI, aligning perfectly with real-world application demands. The core contributions of this work are summarized as follows:
\begin{itemize}
	\item[1)] We propose the SARU framework for RSI shadow detection and removal, which optimizes the linkage between the two sequential stages to mitigate error propagation, forming a synergistic and practical workflow instead of a purely isolated cascaded pipeline.
	\item[2)] We build and open-source two high-quality benchmark datasets (RSISD and SiSRB) for urban RSI shadow analysis, filling the key gap in standardized evaluation data and supporting consistent algorithm comparison for the community.
	\item[3)] We design DBCSF-Net, a dual-branch network that fuses fine-grained spatial boundary details and high-level semantic information, significantly improving shadow detection accuracy and robustness in complex urban scenes.
	\item[4)] We develop the N$^2$SGSR algorithm, a training-free, single-image shadow removal method that operates without paired data, demonstrating strong practical utility and generalization capabilities for real-world RSI applications.
\end{itemize}
\section{Related Work}
\label{sec:rw}
\subsection{Shadow Detection}
Shadow detection is the foundational prerequisite for high-quality removal. Its research has evolved from early physical modeling to modern data-driven deep learning, driven by the need for higher precision in complex urban scenes.

Early traditional shadow detection methods are divided into physical model-driven and handcrafted feature-driven schemes. Physical model-based methods \citep{ADELINE201321, LIASIS2016437} leverage auxiliary data such as solar elevation, azimuth angles, and digital terrain models to predict shadow coverage. While grounded in clear physical principles, these methods are highly dependent on high-quality auxiliary data and struggle to adapt to the complex geometry of urban built environments, limiting their scalability for large-scale high-resolution RSI. Handcrafted feature-based methods extract unique spectral or texture traits of shadows in targeted color spaces (e.g., HSV, Lab) and complete segmentation via thresholding or shallow classifiers \citep{SALVADOR2004238, 4959698}; follow-up studies introduced optimization algorithms like particle swarm optimization to dynamically adjust segmentation thresholds for better adaptability \citep{rs14122756}. Recent works have integrated RSI-specific spectral bands with solar geometry to refine shadow detection \citep{2022EA002387, 10363672}, yet these methods still remain constrained by the inherent limitations of manual feature design and handcrafted physical models. These methods run efficiently but lack robustness in complex, unevenly lit scenes, often misclassifying dark objects as shadows or missing small shadow regions.

Deep learning has reshaped shadow detection thanks to CNNs’ powerful automatic feature extraction capabilities. Early CNN works focused on local feature modeling and boundary refinement:~\cite{7172555} employed stacked CNNs combined with conditional random fields to optimize shadow edges, while~\cite{8578876} incorporated direction-aware contextual encoding to enhance spatial reasoning capabilities. As architectures like U-Net and Transformers matured, end-to-end models emphasizing global-local feature fusion became mainstream \citep{9008769}; some studies also explored GAN-based joint learning of detection and removal, exploring early synergies between the two tasks \citep{8578290}. More recently, large vision models like SAM have been adapted for shadow detection, with task-specific illumination and texture cues injected to improve mask quality \citep{10315174}. While synthetic data has been proposed to alleviate scarcity \citep{9310272}, the distribution gap between synthetic and real RSI shadows remains a core bottleneck. Consequently, bridging the performance gap in complex urban scenarios remains an urgent challenge, necessitating both the development of advanced architectures that effectively couple low-level chromatic variations with high-level semantic context, and the establishment of large-scale, meticulously annotated benchmarks derived from real-world RSI.
\subsection{Shadow Removal}
Shadow removal aims to restore the true spectral information of shadowed pixels, the technical route has transitioned from prior-based restoration to learning-based methods.

Traditional shadow removal approaches have predominantly relied on meticulously designed handcrafted priors, such as illumination estimation, region-based features, shadow density modeling, or user interaction, to construct effective mappings between shadowed and non-shadowed regions. For example,~\cite{5783336} employed Gaussian Mixture Models (GMMs) in conjunction with Markov Random Fields (MRFs) to remove shadows hierarchically. Expanding on spatial consistency,~\cite{7180373} proposed an illumination recovery optimization framework dedicated to accurately estimating the original lighting conditions within shadowed areas,~\cite{6319317} fused surface properties from adjacent non-shadowed regions,~\cite{10.1145/1186415.1186484} modeled shadow density and utilized brightness consistency iteratively to restore occluded regions. For interactive control,~\cite{Gong2016InteractiveRA} introduced a framework where minimal user strokes guide the mapping of brightness and color between shadowed and non-shadowed areas, offering a flexible and precise removal mechanism. While effective for specific scenarios, these methods often suffer from limited generalization and scalability across diverse RSI scenes due to their reliance on strong priors or manual intervention.

With the rapid advancement of deep learning and the increasing availability of dedicated shadow removal datasets, learning-based approaches have achieved substantial breakthroughs. A multitude of GAN-based methods have emerged, where generators are trained to synthesize realistic non-shadow images while discriminators enforce photorealism~\citep{10.1145/3571745, 10154613, 10678116}. These methods often incorporate sophisticated mechanisms such as cycle consistency, residual learning, explicit illumination modeling, and mask constraints to meticulously preserve structural integrity and color consistency. More recently, diffusion models have also garnered significant attention in shadow removal tasks, exploiting latent-space generation and degradation priors to meticulously guide the reconstruction of complex shadow regions~\citep{10376810, 10203768, 10483579}. Beyond these,~\cite{10552804}  introduced a novel neural implicit Fourier transform approach to effectively handle intricate shadows within the frequency domain.~\cite{10191081} proposed a lightweight transformer framework combining a Transformer encoder with a CNN decoder for joint shadow detection and removal. Furthermore, inspired by biological vision,~\cite{10833852} integrated Retinex theory with Mamba networks to successfully achieve robust illumination recovery in RSI. Despite their performance, the critical bottleneck remains their absolute dependency on perfectly paired datasets, which are practically infeasible to collect in real-world RSI scenarios.

To address data scarcity, researchers have explored avenues that reduce reliance on paired ground truth. Some approaches attempt unsupervised learning for style transfer ~\citep{10376810, LIU2024103922, DING2025111354}, they typically assume a relatively consistent scene context (e.g., text or faces), which hinders generalization to the highly diverse and complex characteristics of RSI scenes. To address the data scarcity challenge,~\cite{10641639, 10967107} generated synthetic shadow datasets, but these often involve cropping original images, sacrificing valuable contextual information and failing to capture the full diversity of real-world shadow appearances. To tackle this,~\cite{10.1007/s11263-023-01823-9} proposed an innovative single-image shadow removal framework trained solely with shadow masks and image inputs, employing a lightweight MLP and a histogram-matching discriminator to effectively correct the brightness of shadow regions. Furthermore,~\cite{10678116} designed a self-supervised, single-stage removal network that utilizes dual-branch WGANs to infer region-specific features and perform style alignment with unpaired datasets, offering greater flexibility. In parallel, some superpixel-based segmentation and region-matching methods~\citep{9858933, GUO2024124739, ZHANG2025127769} have demonstrated effectiveness in RSI by leveraging nearby non-shadow areas through histogram alignment or brightness transfer for robust radiometric compensation. With the ascendance of large-scale segmentation models,~\cite{10944212} fine-tuned the SAM model to generate segmentation boundaries with material consistency, subsequently adopting a self-supervised optimization method with color and texture consistency losses to remove shadows effectively. However, a unified framework that inherently integrates detection and a training-free, physically-principled removal without requiring paired data, specifically addressing RSI characteristics, remains an underexplored avenue. This gap is precisely what SARU aims to bridge.

\subsection{Shadow Detection and Removal Datasets}
\label{sec:datasets_rw}
High-quality annotated datasets serve as the foundational cornerstone for deep learning-driven shadow processing. However, within the RS scenes, available benchmarks remain relatively scarce compared to those in the natural image domain. While the AISD dataset \citep{LUO2020443} established an initial large-scale benchmark for urban RSI, its coverage is primarily constrained to specific urban morphologies. Furthermore, although some datasets, such as SSAD \citep{ZHANG2020111945}, those proposed by \citet{SHI2023108557}, and those by \citet{10339831}, provide rich shadow detection scenarios, these resources have not yet been released as open-source, thereby hindering algorithmic validation and progress.

Diverging from these existing works, our proposed RSISD dataset encompasses high-resolution imagery across eleven diverse cities, significantly enhancing spatial and structural diversity to improve model generalization. Crucially, we introduce the SiSRB benchmark, which represents the first dedicated framework for the quantitative evaluation of single-image shadow removal in RSI without relying on temporally paired ground truth. By eliminating the dependency on rare shadow-free references, SiSRB effectively fills a critical gap in standardized assessment for real-world RS applications.

\section{Methodology}
This section details our proposed SARU framework for RSI shadow detection and removal. We begin by outlining the overall architecture in Section~\ref{3.1}. Section~\ref{3.2} elaborates on the shadow detection component, including the Multi-Color Space Combination (MCSC) encoder, DecoupleNet encoder, and Feature Fusion Module (FFM). Section~\ref{3.3} introduces our novel N$^2$SGSR method, designed for fast and realistic shadow removal without pretraining. Finally, Section~\ref{3.4} describes the progressive hierarchical decoder designed to restore spatial resolution and generate the final shadow mask.

\subsection{Framework Description}
\label{3.1}
\begin{figure*}[t]
	\centering
	\includegraphics[width=0.9\textwidth]{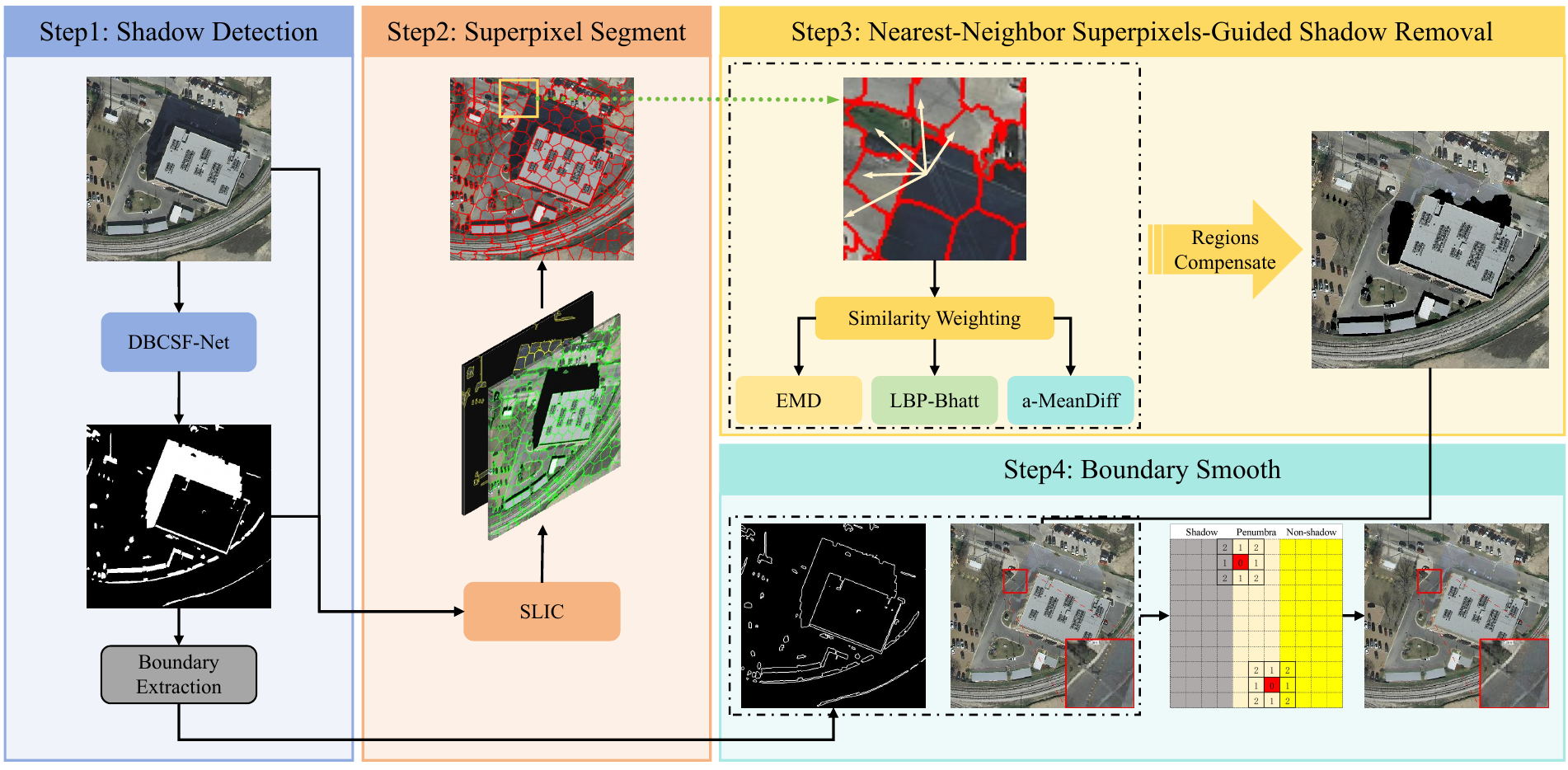}
	\caption{The overall pipeline of the SARU framework. Step 1: Shadow detection via DBCSF-Net; Step 2: Superpixel segmentation using SLIC; Step 3: Shadow removal guided by nearest-neighbor superpixels (N$^2$SGSR); Step 4: Bilateral boundary smoothing for penumbra transition.}
	\label{fig3}
\end{figure*}

\begin{figure*} 
	\centering
	\includegraphics[width=\linewidth]{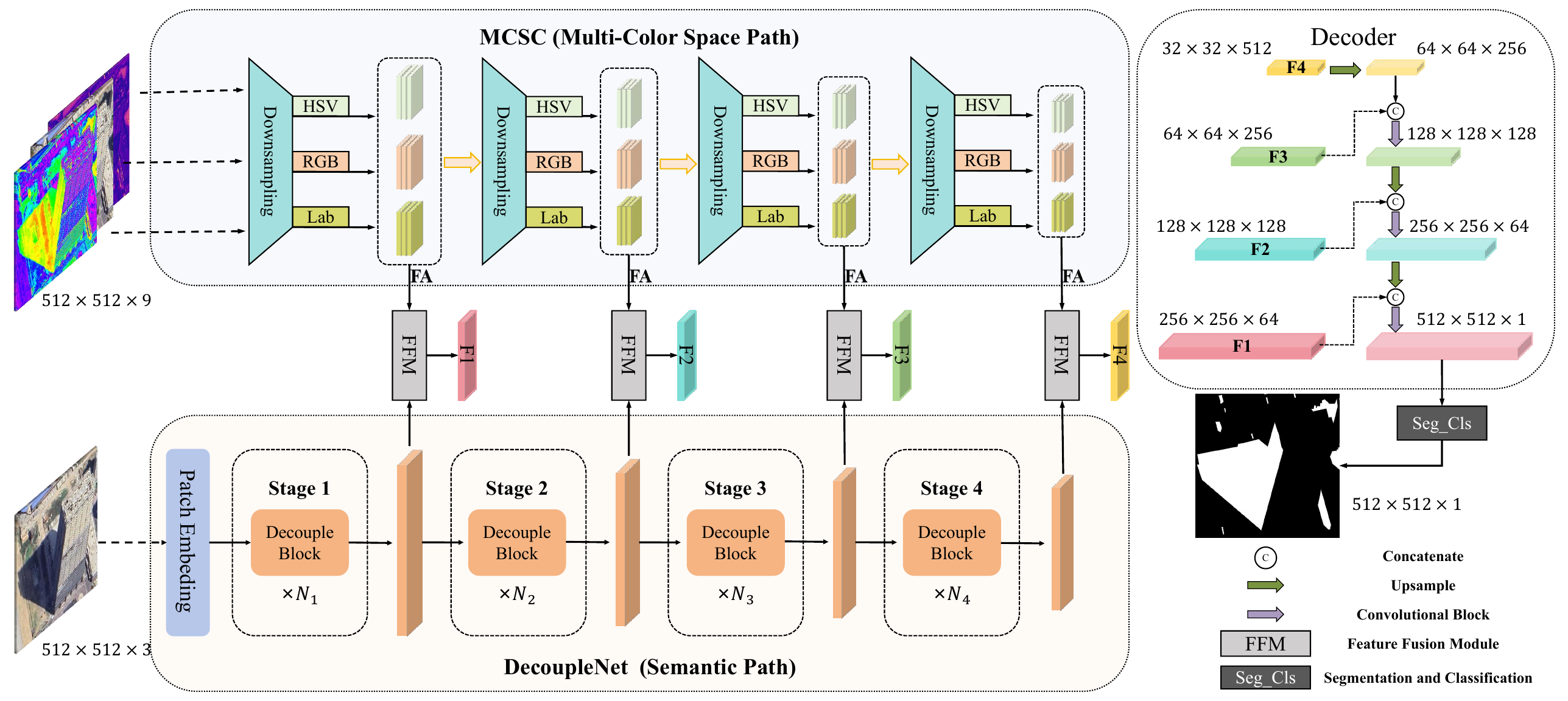}
	\caption{Illustration of the architecture of DBCSF-Net.}
	\label{fig4}
\end{figure*}
Our SARU framework, comprehensively illustrated in Fig.~\ref{fig3}, represents a novel paradigm that seamlessly integrates shadow detection and illumination-consistent removal into a unified pipeline. The process begins with our pre-trained DBCSF-Net, which meticulously predicts a high-fidelity shadow mask from the input shadowed image. Combined with the original image, it directs superpixel segmentation using SLIC, performed independently on shadow and non-shadow regions, which are then judiciously concatenated. Each shadow superpixel is then processed sequentially by the N$^2$SGSR algorithm. The method quantifies the contribution weights of reference superpixels toward shadow illumination compensation by evaluating their similarity to shadowed regions through Earth Mover's Distance (EMD) \citep{Rubner2000}, Local Binary Pattern-Bhattacharyya (LBP) \citep{1017623}, and a-channel Mean Difference (a-MeansDiff) metrics. To further mitigate artifacts and ensure smooth transitions, particularly in penumbra regions, a dedicated bilateral boundary smoothing step is applied at the final stage.

\subsection{Shadow Detection}
\label{3.2}
Accurate shadow detection is paramount for effective shadow removal. As illustrated in Fig.~\ref{fig4}, DBCSF-Net comprises a MCSC encoder and a DecoupleNet encoder, which collaboratively extract comprehensive shadow-related features. These features are then fused and refined by a FFM to generate precise shadow masks.

\subsubsection{MCSC Encoder}
To robustly capture fine-grained chromatic and spatial information under diverse illumination conditions, we propose the MCSC Encoder. This module employs parallel encoding branches across RGB, HSV, and Lab color spaces, as depicted in the multi-color space path of Fig.~\ref{fig4}. Given an input RGB image $I_{\text{RGB}}$, it is transformed into HSV and Lab spaces:
\begin{equation}
	I_{\text{RGB}} \rightarrow \left\{ I_{\text{HSV}}, I_{\text{Lab}} \right\},
\end{equation}
where $I_{\text{HSV}}$ and $I_{\text{Lab}}$ represent the respective color space representation. Each color space input is then processed by an independent $\mathrm{ConvBlock}$ to generate feature maps $F_{\text{sp}}$, where $\text{sp} \in \{\text{RGB}, \text{HSV}, \text{Lab}\}$:
\begin{equation}
	F_{\text{sp}} = \mathrm{ConvBlock} \left( I_{\text{sp}} \right), \quad F_{\text{sp}} \in \mathbb{R}^{H' \times W' \times C'},
\end{equation}
where $ H' $, $ W' $ is spatial dimensions and $ C' $ is channel count.

\label{3.2.1}
To effectively integrate these multi-color-space features, a cross-spatial attention mechanism is introduced. This mechanism generates attention weights $\omega_{\text{sp}}$ by concatenating the feature maps, applying global average pooling, and passing them through fully connected layers:
\begin{equation}
	\omega_{\text{sp}} = \mathrm{Attention} \left( \mathrm{Concat} \left( F_{\text{RGB}}, F_{\text{HSV}}, F_{\text{Lab}} \right) \right),
\end{equation}
the final color feature $F_{\text{color}}$ is obtained through a weighted summation:
\begin{equation}
	F_{\text{color}} = \sum_{\text{sp}} \omega_{\text{sp}} \odot F_{\text{sp}}.
\end{equation}

\subsubsection{DecoupleNet Encoder}
\label{3.2.2}
Distinguishing shadows from semantically similar dark objects (e.g., black buildings) in RSI is a significant challenge. To address this, we incorporate a semantic information extraction branch using the lightweight DecoupleNet~\citep{10685518}, shown in the semantic path of Fig.~\ref{fig4}. This encoder efficiently decouples spatial and semantic features, reducing computational complexity while maintaining high semantic segmentation accuracy. The input image $ I $ is first processed by a Patch Embedding module, which partitions it into non-overlapping patches and extract initial features:
\begin{equation}
	F_{\text{init}} = \mathrm{PatchEmbed}\left(I\right), \quad F_{\text{init}} \in \mathbb{R}^{H' \times W' \times C'}.
\end{equation}

Subsequently, multi-scale semantic feature maps $F_i^{\text{sm}}$ are generated across four stages. Each stage comprises feature decoupling and integration modules within an MLPblock, capturing multi-scale semantic information and adjusting feature resolution and channels:
\begin{equation}
	F_i^{\text{sm}} = \mathrm{MLPblock}_i \left(F_{i-1}^{\text{sm}}\right), \quad F_0^{\text{sm}} = F_{\text{init}}.
\end{equation}

\begin{figure}
	\raggedleft      
	\includegraphics[width=\linewidth]{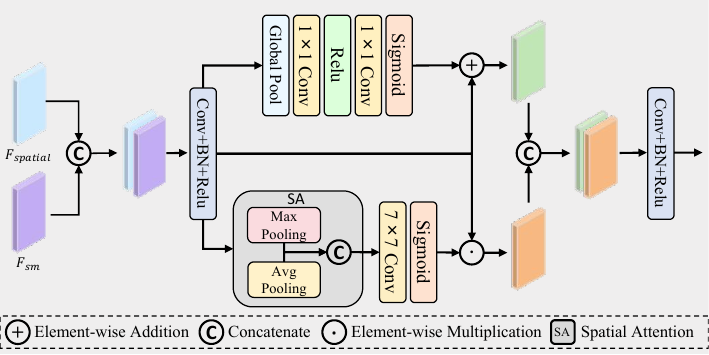}
	\caption{Structural diagram of FFM.}
	\label{fig5}
\end{figure}

\subsubsection{FFM}
\label{3.2.3}
To optimally combine the color features $F_{\text{color}}$ from the MCSC encoder and the semantic features $F_{\text{sm}}$ from the DecoupleNet encoder, we designed the FFM, depicted in Fig.~\ref{fig5}. FFM leverages channel and spatial attention mechanisms to enhance feature complementarity, yielding high-quality fused features for the shadow detection task.

The FFM workflow begins by concatenating $F_{\text{color}}$ and $F_{\text{sm}}$ and processing them with a $1 \times 1$ convolution to generate initial fused features. A channel attention branch then computes channel weights via global pooling, emphasizing critical channel information. Concurrently, a spatial attention branch generates attention maps using average and max pooling to highlight significant spatial regions. Both branches incorporate residual connections to optimize feature representation. Finally, the outputs of these two branches are concatenated and integrated with a $1 \times 1$ convolution to produce the final fused feature $F_i^{\text{fuse}} (H' \times W' \times C')$, which serves as a skip connection for the subsequent decoder. The fused feature $F_i^{\text{fuse}}$ is ultimately upsampled by a deconvolution-based decoder to predict a binary shadow mask, ensuring precise shadow delineation with enhanced boundary integrity and reduced false positives.

\subsection{Decoder Structure}
\label{3.4}
To restore spatial resolution and accurately reconstruct complex shadow boundaries, we design a progressive hierarchical decoder, as illustrated in the right part of Fig. \ref{fig4}. This decoder integrates high-level semantic information with low-level spatial details through transposed convolutions and skip connections.

The decoder takes four multi-scale features $\left\{F_1^{\text{fuse}}, F_2^{\text{fuse}}, F_3^{\text{fuse}}, F_4^{\text{fuse}}\right\}$ from FFM as input. The reconstruction process starts from the deepest feature $F_4^{\text{fuse}}$. To mitigate the information loss inherent in deep convolutional networks, encoder features of the same spatial resolution are concatenated along the channel dimension at each stage. Each decoding layer is formally defined as follows:
\begin{equation}
	F_i^{\text{rs}} = 
	\begin{cases}
		\mathrm{Up}\left(F_4^{\text{fuse}}\right), & i=1 \\
		\mathrm{RfConv}\left(\mathrm{Cat}\left(\mathrm{Up}\left(F_{i-1}^{\text{rs}}\right), F_{4-i}^{\text{fuse}}\right)\right), & i > 1
	\end{cases},
\end{equation}
where $\mathrm{Up}(\cdot)$ denotes a $4 \times 4$ transposed convolution with a stride of 2. The refinement module $\mathrm{RfConv}(\cdot)$ consists of two consecutive blocks of $3 \times 3$ convolution, Batch Normalization (BN), and ReLU activation.

After three stages of progressive refinement, the feature maps are restored to the original input resolution. Finally, a $1 \times 1$ convolutional layer followed by a Sigmoid activation function is employed to generate the final shadow mask $I_{\text{m}}$:
\begin{equation}
	I_{\text{m}} = \sigma\left(\mathrm{Conv}_{1 \times 1}\left(F_4^{\text{rs}}\right)\right),
\end{equation}
where $\sigma(\cdot)$ represents the Sigmoid function.

\subsection{Shadow Removal}
\label{3.3}
After obtaining the shadow mask, to remove shadows within the mask, we introduce N$^2$SGSR. This method is based on the camera imaging principle, where a pixel's intensity value $ I_x $ is the product of its illumination intensity $ L_x $ and its reflectance $ R_x $:
\begin{equation}
	I_x = L_x \cdot R_x.
\end{equation}

The illumination intensity $L_x$ comprises ambient light $L_x^{\text{a}}$ and direct light $L_x^{\text{d}}$. Pixels in different regions can be expressed as:
\begin{equation}
	\begin{cases}
		I_x^{\text{s}} = L_x^{\text{a}} \cdot R_x, \\
		I_x^{\text{ns}} = \left(L_x^{\text{d}} + L_x^{\text{a}}\right) \cdot R_x, \\
		I_x^{\text{ps}} = \left(\alpha_x L_x^{\text{d}} + L_x^{\text{a}}\right) \cdot R_x, \quad \alpha_x \in \left(0, 1\right),
	\end{cases}
\end{equation}
where $I_x^{\text{s}}$, $I_x^{\text{ns}}$, and $I_x^{\text{ps}}$ denote pixel values in umbra, non-shadow, and penumbra regions, respectively, with $\alpha_x$ as the illumination degradation factor.

The ratio of direct to ambient light intensity, $r_x$, for any pixel $x$ can be approximated as:
\begin{equation}
	\begin{split}
		r_x &= \frac{I_x^{\text{ns}} - I_x^{\text{s}}}{I_x^{\text{s}}} = \frac{\left[\left(L_x^{\text{d}} + L_x^{\text{a}}\right) - \left(\alpha_x L_x^{\text{d}} + L_x^{\text{a}}\right)\right] \cdot R_x}{\left(\alpha_x L_x^{\text{d}} + L_x^{\text{a}}\right) \cdot R_x}\\ &= \frac{\left(1 - \alpha_x\right) \cdot L_x^{\text{d}}}{\alpha_x L_x^{\text{d}} + L_x^{\text{a}}}
	\end{split}.
\end{equation}

For a fully shadowed region ($\alpha_x = 0$), the re-illuminated pixel value $I_x^{\text{sr}}$ simplifies to:
\begin{equation}
	I_x^{\text{sr}} = \left(r_x + 1\right) \cdot I_x^{\text{s}}.
\end{equation}

The core principle of shadow removal lies in the fact that, under the same background and material composition, shadow regions and adjacent non-shadow regions typically exhibit remarkably similar surface reflectance properties. This fundamental characteristic allows us to leverage neighboring non-shadow superpixels as a reliable reference to restore the true appearance and illumination of shadowed areas. Specifically, for shadowed superpixels in an image that have adjacent non-shadowed regions, we construct a distance matrix based on Euclidean distance to precisely locate their $n$ nearest non-shadowed superpixels as local reference points. However, in certain extreme cases, such as when a large homogeneous region is completely covered by shadows, and the background material of adjacent non-shadow superpixels does not match that of the shadow superpixels, their similarity weights will all be lower than the set threshold of 0.2. The framework will interpret this as material mismatch, expand the search range, conduct a similarity-weighted global search, and ensure the material perception capability of illumination compensation under the constraints of Equation(\ref{eq19}).

To effectively utilize information from matched non-shadow superpixels, we first perform a preliminary shadow removal based on global non-shadow region statistics. Subsequently, we systematically quantify the contribution weights of non-shadow superpixels using three distinct metrics:

\textbf{EMD in Lab Color Space:} Shadowing alters an object's color distribution in the Lab color space. EMD quantifies the color distribution difference between shadow and non-shadow superpixels, where a smaller difference indicates a greater contribution. The EMD for each channel is computed:
\begin{equation}
	\begin{cases}
		d_{\text{L}} = \mathcal{W} \left( L_{\text{s}} ; L_{\text{ns}} \right) \\
		d_{\text{a}} = \mathcal{W} \left( a_{\text{s}} ; a_{\text{ns}} \right) \\
		d_{\text{b}} = \mathcal{W} \left( b_{\text{s}} ; b_{\text{ns}} \right)
	\end{cases},
\end{equation}
where $\mathcal{W}\left(\cdot\right)$ denotes the Wasserstein distance \citep{Panaretos_2019} between two distributions, defined as:
\begin{equation}
	\mathcal{W}_1\left(\mu,\nu\right) =
	\inf_{\gamma\in\Gamma\left(\mu,\nu\right)}\int_{\mathcal{X}\times\mathcal{X}}\left\|x-y\right\| \,\mathrm{d}\gamma\left(x,y\right),
\end{equation}
where $\Gamma\left(\mu, \nu\right)$ represents the set of all coupling distributions between $\mu$ and $\nu$, and $\|\cdot\|$ denotes the Euclidean norm.
The combined EMD distance is:
\begin{equation}
	D_{\text{EMD}} = \frac{d_{\text{L}} + d_{\text{a}} + d_{\text{b}}}{300},
\end{equation}
where $L_{\text{s}}, a_{\text{s}}, b_{\text{s}}$ and $L_{\text{ns}}, a_{\text{ns}}, b_{\text{ns}}$ are channel values for shadow and non-shadow superpixels, respectively.

\textbf{LBP Texture Similarity:} Texture information is generally robust to illumination changes. We convert images to Local Binary Pattern (LBP) grayscale maps and compute texture similarity using the Bhattacharyya coefficient:
\begin{equation}
	\rho_{\text{LBP}} = \sum_{x=1}^M \sqrt{H_{\text{s}}\left(x\right) \cdot H_{\text{ns}}\left(x\right)},
\end{equation}
with the distance metric:
\begin{equation}
	D_{\text{LBP}} = 1 - \rho_{\text{LBP}},
\end{equation}
where $H_{\text{s}}\left(x\right)$ and $H_{\text{ns}}\left(x\right)$ are the probabilities of the two regions’ LBP histograms at the $x$-th bin. Higher similarity implies greater contribution.

\textbf{$a$-Channel Mean Difference:} The $a$-channel in Lab color space exhibits minimal attenuation in shadow regions. Thus, a smaller difference in $a$-channel mean values between superpixels indicates higher similarity and greater contribution.
\begin{equation}
	D_{\text{a-means}} = \frac{| \mu_a^{\text{s}} - \mu_a^{\text{ns}} |}{128},
\end{equation}
where $\mu_a^{\text{s}}$ and $\mu_a^{\text{ns}}$ are the mean $a$-channel values.

Based on this theoretical framework, we propose a systematic two-level priority-aware mechanism for umbra shadow removal:
\begin{enumerate}
	\item[1)] Initially, we apply Simple Linear Iterative Clustering (SLIC) for superpixel segmentation on both shadow and non-shadow regions, guided by the shadow mask $I_{\text{m}}$. These segments are subsequently merged to form the final segmented image $I_{\text{sp}}$. Based on the shadow mask, each superpixel is then classified: if the shadow proportion within the region exceeds 80\%, it is labeled as a shadow superpixel. For each superpixel, we extract Lab color space features $I_{\text{Lab}}$ through color space transformation and texture features $I_{\text{LBP}}$ using the LBP algorithm to capture local structural properties.
	
	\item[2)] For every shadow superpixel $SP_x^{\text{s}}$, we identify a candidate set of $n$ nearest non-shadow superpixels, denoted as $\{SP_1^{\text{near}}, \ldots, SP_n^{\text{near}}\}$. This spatial proximity is determined by a precomputed Euclidean distance matrix, ensuring that the illumination references are drawn from the most geographically relevant areas.
	
	\item[3)] To prevent feature contamination (e.g., incorrectly restoring road shadows using building roof properties), we establish a selection priority by calculating the contribution weight $\omega_i^{\text{near}}$ for each candidate. This weight is evaluated using three semantic-consistency metrics: $D_i^{\text{EMD}}$ (color distribution), $D_i^{\text{LBP}}$ (texture similarity), and $D_i^{\text{a-means}}$ (spectral attenuation):
	\begin{equation}
		\label{eq19}
		\omega_i^{\text{near}} = \frac{1}{\alpha D_i^{\text{EMD}} + \beta D_i^{\text{LBP}} + \gamma D_i^{\text{a-means}} + \varepsilon},
	\end{equation}
	where $\alpha, \beta, \gamma$ are hyperparameters that penalize candidates with mismatched materials, and $\varepsilon = 10^{-4}$ is a stability constant. This mechanism ensures that non-shadow regions with identical semantic properties receive the highest priority.
	
	\item[4)] For each shadow superpixel $SP_x^{\text{s}}$ and its adjacent non-shadow superpixel $SP_i^{\text{near}}$, the pixel value ratio is calculated as:
	\begin{equation}
		r_i^x = \frac{I_i^{\text{ns}} - I_x^{\text{s}}}{I_x^{\text{s}}},
	\end{equation}
	where $I_x^{\text{s}}$ and $I_i^{\text{ns}}$ represent the pixel values of the shadow and non-shadow superpixels, respectively. The aggregated ratio $r_x$ for $SP_x^{\text{s}}$ is then computed as a weighted sum:
	\begin{equation}
		r_x = \sum_{i=1}^n r_i^x \cdot \omega_i^{\text{near}}.
	\end{equation}
	
	\item[5)] The reconstructed pixel value $I_x^{\text{sr}}$ for the umbra region is derived using the computed ratio $r_x$ and the physical illumination model:
	\begin{equation}
		\label{eq20}
		I_x^{\text{sr}} = \left(r_x + 1\right) \cdot I_x^{\text{s}}.
	\end{equation}
	
	\item[6)] Steps 2) to 5) are iterated until all shadow superpixels are processed. Notably, the restoration of each shadow unit is independent, as it draws illumination references exclusively from the original non-shadow regions. This order-independence ensures that the final result is robust to the processing sequence and effectively mitigates the propagation of reconstruction errors.
	
\end{enumerate}

Direct application of Equation(\ref{eq20}) may lead to artifacts or discontinuities in penumbra regions, which are transitional zones. To address this, we introduce a bilateral boundary processing method. We first extract the penumbra region from the shadow mask $ M_{\text{ps}} $ using dilation and erosion operations:
\begin{equation}
	M_{\text{ps}} = \mathrm{Dilate}\left(M\right) - \mathrm{Erode}\left(M\right),
\end{equation}
where $\mathrm{Dilate}\left(M\right)$ and $\mathrm{Erode}\left(M\right)$ denote dilation and erosion with a radius of $ r $. 

A bilateral boundary filter is then applied, using Manhattan distance as a weighting basis, to perform inward and outward diffusion-based smoothing. This ensures improved boundary smoothness and consistency across the penumbra, significantly enhancing the visual quality of the restored images and preventing unnatural edges, as conceptually depicted in Fig.~\ref{fig3}.

\section{Proposed Dataset}
\begin{figure}
	\centering \includegraphics[width=1\linewidth]{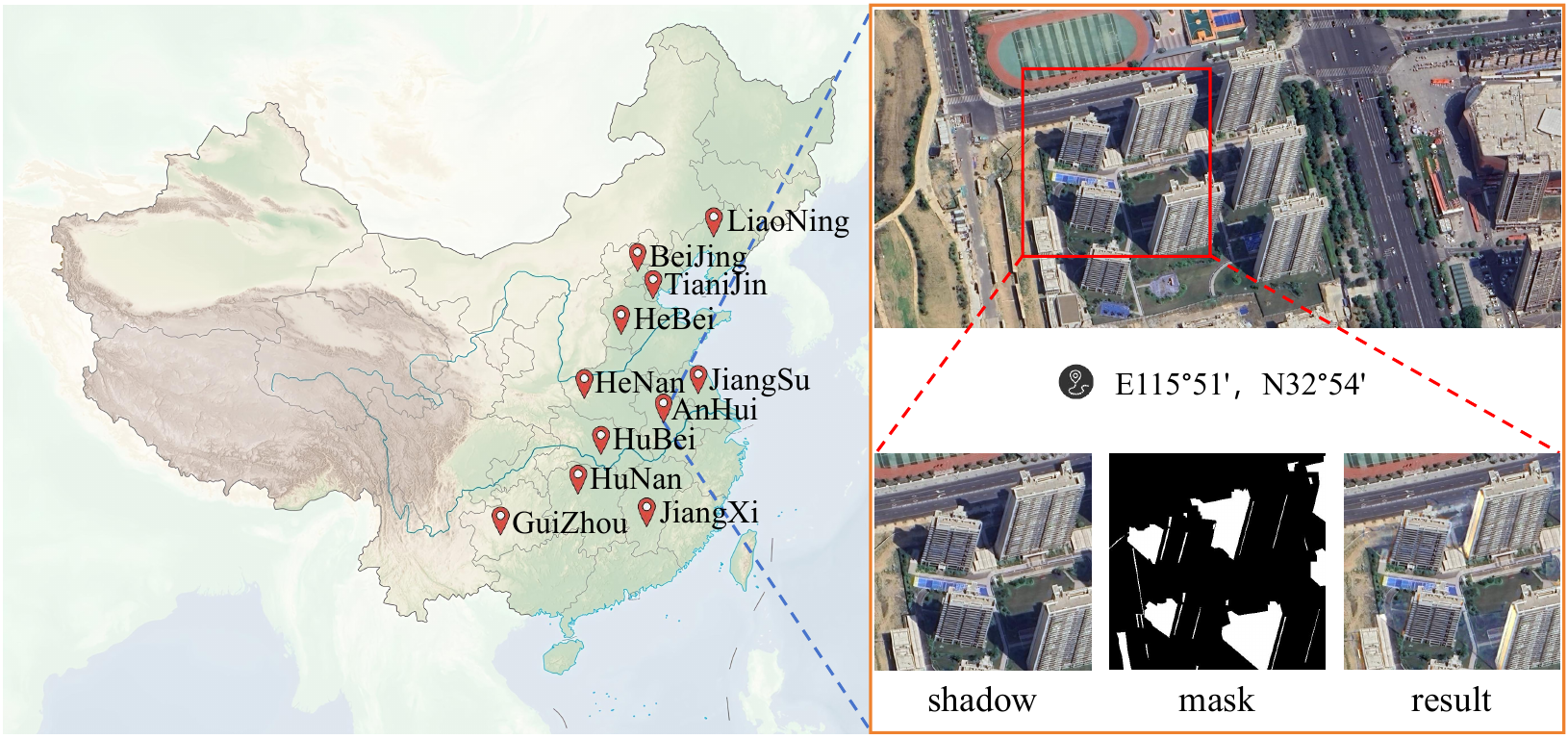} 
	\caption{The left image illustrates the eleven different cities in China covered by the RSISD dataset. The right image presents a sampling site, which includes the original shadow image, the manually annotated shadow mask, and the shadow removal result generated by our proposed method.}
	\label{sample} 
\end{figure}

The RSISD dataset was specifically curated to address the critical scarcity of dedicated, high-quality datasets for RSI shadow detection tasks. As depicted in Fig.~\ref{sample}, we meticulously annotated both cast shadows and self-shadows within the images, ensuring precise and comprehensive ground truth for shadow detection research. The images were sampled from eleven different cities across China, each possessing a spatial resolution of $0.3 m$ and a fixed size of $512 \times 512$ pixels. For fair evaluation and reproducibility, the dataset is meticulously divided into training, validation, and test sets. The RSISD dataset encompasses a diverse range of urban scenarios, including factories, residential areas, schools, and stadiums, offering a robust platform for method development and assessment, particularly for the challenging task of distinguishing shadows from dark objects in complex urban environments.

The SiSRB dataset was constructed to address the pivotal challenge of evaluating single-image shadow removal without paired non-shadow data. This was achieved by randomly selecting 55 shadow images from the test and validation sets of both the AISD and RSISD datasets. Each image includes manually annotated evaluation masks delineating shadow and non-shadow regions within the same background, with visually exemplified in Fig.~\ref{fig6}. This dataset is the first of its kind specifically designed to rigorously benchmark single-image shadow removal performance in RSI without reliance on paired ground truth, fostering more practical and generalizable research.

\section{Experiments}
This section first delineates the experimental setup, detailing the datasets utilized and the evaluation protocols applied, thereby ensuring the reproducibility and rigor of the testing scenarios. Subsequently, it presents the empirical results of our proposed DBCSF-Net on two benchmark RSI shadow detection datasets. Following this, the performance of our N$^2$SGSR method in the context of RSI shadow removal is thoroughly evaluated. Qualitative visualizations are then provided to offer intuitive demonstrations of our method's advantages over existing state-of-the-art approaches.Finally, a series of comprehensive ablation studies are conducted to systematically dissect and quantify the individual contributions of each architectural component to the overall efficacy of the proposed methodology. 

\subsection{Experimental Settings}
For the training of the DBCSF-Net, the model was implemented using the PyTorch framework and was exclusively trained on an NVIDIA RTX $3090$ GPU. Optimization was achieved with the Adam optimizer, employing a consistent learning rate of $ 2 \times 10^{-4} $. The batch size was set to $4$, and the training process spanned $30$ epochs. Regarding the shadow removal phase using N$^2$SGSR, The number of pixels per superpixel was set to $600$, and the number of neighboring superpixels was fixed at $7$, while the critical parameters $\alpha$, $\beta$, and $\gamma$ were determined empirically to be $0.6$, $0.3$, and $0.1$, respectively. These settings were chosen to ensure optimal performance and comparability across experiments.

\subsection{Datasets}
Our experiments on shadow detection and removal were rigorously conducted on two established benchmark datasets, supplemented by a newly constructed dataset for specific shadow removal evaluation.
\begin{enumerate}
	\item The AISD~\citep{LUO2020443} dataset comprises $514$ image pairs, each consisting of a shadow image and its corresponding binary mask. This dataset was partitioned into $412$ pairs for training, $51$ pairs for validation, and $51$ pairs for testing.
	\item Our proposed RSISD dataset encompasses $1000$ training pairs, $86$ validation pairs, and $200$ testing pairs, offering a substantial collection for robust model training and evaluation.
	\item Our proposed SiSRB dataset comprises 55 shadow images, each accompanied by a corresponding evaluation mask, with distinct colors used to differentiate the regions to be evaluated.
\end{enumerate}

\begin{figure}
	\centering
	\includegraphics[width=\linewidth]{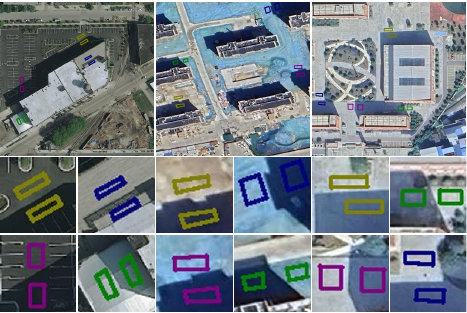}
	\caption{For the selected images, we annotate shadowed and non-shadowed areas of similar land cover types, ensuring that the number of pixels within pairs of annotated areas is roughly balanced, thus providing a fair basis for comparative analysis.}
	\label{fig6}
	\vspace{-2mm}
\end{figure}

\subsection{Evaluation Indices}
For the shadow detection task, our method was benchmarked against $11$ state-of-the-art approaches: RSiSD~\citep{SILVA2018104}, AFFPN~\citep{9242301}, ECA-SD~\citep{10.1145/3474085.3475199}, FDRNet~\citep{9711042}, SDCM~\citep{10.1145/3503161.3547904}, RSD~\citep{WU2023110614}, SDDNet~\citep{Cong2023SDDNetSD}, SILT~\citep{10377292}, CADDN~\citep{fernandez2024shadow}, SACNet~\citep{Chen2025}, and SDKU-Net~\citep{computers14030080}. To ensure a fair and consistent comparison, all methods, with the exception of SACNet and SDKU-Net whose codes were not publicly available, were re-trained and re-evaluated on our designated hardware infrastructure using their respective official implementations. For SACNet and SDKU-Net, the performance metrics reported in their original publications were directly adopted. Consistent with prior research in shadow detection, we employed four widely recognized evaluation metrics: Recall, F1-score, Balanced Error Rate (BER), and Intersection over Union (IoU).

For the shadow removal task, we compared against $7$ state-of-the-art approaches: RSiSR~\citep{SILVA2018104}, G2R-ShadowNet~\citep{9578655}, MEF-SR~\citep{9874905}, SRMF~\citep{9858933}, Self-ShadowGAN~\citep{10.1007/s11263-023-01823-9}, MAOSD~\citep{ZHANG2025127769}, and RS-GSSR~\citep{10967107}. Analogous to the shadow detection experiments, all comparative methods were re-trained and assessed within the same computational environment using their authors' provided source codes. Considering the inherent challenge of evaluating shadow removal in the absence of temporally paired ground truth, we adopt the widely-used non-reference metrics NIQE and BRISQUE \citep{6272356}, and further design a dedicated material consistency evaluation protocol based on the SiSRB benchmark. As illustrated in Fig.~\ref{fig6}, we compute the Shadow Recovery Index (SRI) and Color Dissimilarity (CD) by comparing shadow regions with their adjacent well-lit reference counterparts, which enables a reliable and quantitative measurement of the illumination restoration fidelity.

\begin{table*}[t]
	\centering \scriptsize
	\caption{Comparison of shadow detection methods on AISD and our proposed RSISD datasets. $\uparrow$ indicates higher is better, $\downarrow$ indicates lower is better. Best results are in \textbf{bold}, second-best are \underline{underlined}.}
	\renewcommand{\arraystretch}{1.0}
	\setlength{\tabcolsep}{7pt}
	\begin{tabular}{l c cccc cccc} 
		\toprule
		\multirow{2.5}{*}{\centering Method} & \multirow{2.5}{*}{\centering Publication} & \multicolumn{4}{c}{AISD} & \multicolumn{4}{c}{RSISD} \\
		\cmidrule(r){3-6} \cmidrule(r){7-10}
		& & Accuracy$\uparrow$ & $F_1\text{-score}$$\uparrow$ & BER$\downarrow$ & IoU$\uparrow$ & Accuracy$\uparrow$ & $F_1\text{-score}$$\uparrow$ & BER$\downarrow$ & IoU$\uparrow$ \\
		\midrule
		RSiSD~\citep{SILVA2018104} & ISPRS & 85.70 & 54.56 & 30.57 & 38.70 & 83.38 & 53.70 & 30.65 & 28.20 \\
		AFFPN~\citep{9242301} & SPL & \underline{96.84} & 92.15 & 5.22 & \underline{85.52} & 96.76 & 92.54 & 5.44 & 86.38 \\
		ECA-SD~\citep{10.1145/3474085.3475199} & ACM & \textbf{98.49} & 85.52 & 11.70 & 76.20 & \underline{97.56} & 90.33 & 6.64 & 84.40 \\
		FDRNet~\citep{9711042} & ICCV & 93.62 & 86.23 & 5.85 & 75.97 & 95.36 & 90.24 & 4.72 & 82.58 \\
		SDCM~\citep{10.1145/3503161.3547904} & ACM & 94.25 & 87.87 & 4.79 & 78.51 & 97.36 & \underline{94.01} & 3.89 & \underline{88.89} \\
		RSD~\citep{WU2023110614} & KBS & 96.07 & 90.79 & 4.48 & 83.25 & 96.79 & 92.22 & 4.79 & 86.11 \\
		SDDNet~\citep{Cong2023SDDNetSD} & ACM & 95.31 & 89.71 & \textbf{4.24} & 81.43 & 96.60 & 92.32 & 3.91 & 86.10 \\
		SILT~\citep{10377292} & ICCV & 94.11 & 87.65 & \underline{4.47} & 78.16 & 96.37 & 92.28 & \textbf{3.26} & 85.92 \\
		CADDN~\citep{fernandez2024shadow} & IVC & 95.97 & 90.55 & 5.91 & 82.86 & 92.07 & 77.33 & 13.94 & 68.47 \\
		SACNet~\citep{Chen2025} & JGSA & -- & \underline{92.27} & 5.02 & 81.51 & -- & -- & -- & -- \\
		SDKU-Net~\citep{computers14030080} & Computers & 96.50 & 92.19 & 5.08 & \underline{85.52} & -- & -- & -- & -- \\
		\midrule
		\rowcolor[gray]{0.95} DBCSF-Net(Ours) & -- & 96.79 & \textbf{92.65} & 4.69 & \textbf{86.35} & \textbf{97.59} & \textbf{94.48} & \underline{3.75} & \textbf{89.71} \\
		\bottomrule
	\end{tabular}
	\label{tab:shadow_detection_comparison}
\end{table*}

\begin{table*}[t]
	\centering \scriptsize
	\caption{Quantitative comparison of shadow removal methods on the AISD and SiSRB datasets. $\uparrow$ indicates higher is better, $\downarrow$ indicates lower is better. The best, second-best, and third-best results are highlighted in \textbf{bold}, \underline{underlined}, and \textit{italic}, respectively.}
	\label{tab:shadow_removal_comparison}
	\renewcommand{\arraystretch}{1.1}
	\setlength{\tabcolsep}{4pt} 
	\begin{tabular}{l c ccccc ccccc}
		\toprule
		\multirow{2.5}{*}{Method} & \multirow{2.5}{*}{Publication} & \multicolumn{5}{c}{AISD} & \multicolumn{5}{c}{SiSRB} \\
		\cmidrule(lr){3-7} \cmidrule(lr){8-12}
		& & SRI$\uparrow$ & CD$\downarrow$ & NIQE\_S$\downarrow$ & BRISQUE$\downarrow$ & Times$\downarrow$ & SRI$\uparrow$ & CD$\downarrow$ & NIQE\_S$\downarrow$ & BRISQUE$\downarrow$ & Times$\downarrow$ \\
		\midrule
		Shadow Image & -- & 0.39 & 37.06 & 2.11 & 17.89 & -- & 0.38 & 42.86 & 1.97 & 27.88 & -- \\
		RSiSR~\citep{SILVA2018104} & ISPRS & 0.65 & 23.52 & 1.50 & 18.95 & \textit{5.37s} & 0.53 & 35.15 & \textit{1.37} & 28.17 & \textit{5.26s} \\
		G2R-ShadowNet~\citep{9578655} & CVPR & 0.80 & 13.95 & 1.40 & \textit{14.11} & -- & 0.83 & 15.25 & 2.01 & 23.41 & -- \\
		MEF-SR~\citep{9874905} & TGRS & 0.49 & 31.78 & 1.82 & 14.18 & \underline{1.53s} & 0.46 & 38.15 & 1.65 & 24.29 & \underline{1.21s} \\
		SRMF~\citep{9858933} & ICME & \underline{0.88} & 10.52 & 1.19 & 15.15 & 90.35 & \underline{0.88} & 11.38 & 1.46 & 24.08 & 64.51 \\
		Self-ShadowGAN~\citep{10.1007/s11263-023-01823-9} & IJCV & 0.85 & 10.02 & 1.28 & 18.07 & 256s & 0.76 & 20.69 & \textbf{1.33} & 25.01 & 243s \\
		MAOSD~\citep{ZHANG2025127769} & ESWA & 0.86 & \underline{9.57} & \textbf{1.09} & 17.78 & 14.99s & \textit{0.87} & \underline{9.93} & \underline{1.34} & 26.82 & 14.63s \\
		RS-GSSR~\citep{10967107} & TGRS & \textbf{0.90} & \textbf{6.81} & \textit{1.16} & \textbf{11.42} & -- & \textbf{0.89} & \textbf{9.82} & 1.45 & \textit{23.01} & -- \\
		\midrule
		\rowcolor[gray]{0.95} \textbf{N$^2$SGSR (Ours)} & -- & \textit{0.87} & \textit{9.80} & \underline{1.13} & \underline{12.14} & \textbf{1.39s} & \textbf{0.89} & \textit{10.61} & \textbf{1.33} & \textbf{21.46} & \textbf{1.24s} \\
		\bottomrule
	\end{tabular}
\end{table*}

\subsection{Experimental Results}
\subsubsection{Results on Shadow Detection}

Table~\ref{tab:shadow_detection_comparison} presents a comprehensive comparison of the proposed method's performance against various state-of-the-art shadow detection approaches on both the AISD and RSISD datasets. The results unequivocally demonstrate that our method consistently achieved either the best or second-best performance across the majority of the evaluated metrics, particularly the $F_1\text{-score}$ and IoU, which are critical indicators of detection quality. This robust outcome substantiates that the judiciously designed DBCSFM, by effectively integrating multi-color space features with rich semantic information, is highly capable of more accurately discerning dark objects and delineating shadow regions. Crucially, it achieves this while meticulously preserving boundary integrity and detail continuity, which are vital for high-quality RSI analysis.

However, a closer inspection of the AISD dataset reveals that our method did not attain the absolute best BER metric when compared to some existing approaches. This slight discrepancy is hypothesized to stem from the substantial presence of vegetation shadows within the AISD dataset, which often exhibit irregular textures and subtle boundaries. The annotations for these vegetation shadows in AISD are often of relatively lower quality and consistency, which inherently complicates the model's learning process. Consequently, this can lead to incomplete detection of subtle or semi-transparent shadows, potentially increasing false positives or false negatives in these specific areas. Furthermore, the interplay of uneven local illumination and intricate building structures in such scenes can introduce false positives, thereby negatively impacting the overall error rate. Nevertheless, through the strategic incorporation of attention mechanisms, our method effectively amplified the feature contrast between shadow and non-shadow regions. This significantly mitigated common issues such as shadow fragmentation, blurred edges, and the challenge of detecting small shadows. Ultimately, our approach exhibited strong generalization capability even under these demanding conditions, showcasing its robustness across diverse urban RSI scenarios.

\subsubsection{Results on Shadow Removal}
The quantitative performance of the shadow removal task is summarized in Table~\ref{tab:shadow_removal_comparison}. Our N$^2$SGSR demonstrates an exceptional balance between restoration fidelity and computational efficiency across both the AISD and SiSRB datasets. Compared to traditional detection-to-removal frameworks, such as RSiSR and MEF-SR, our approach consistently yields superior results in terms of SRI and CD metrics. This enhancement is primarily attributable to its more accurate and robust shadow detection capabilities in the initial stage, which, in turn, translates into significantly improved shadow removal outcomes. Notably, in the context of no-reference image quality assessment, our method achieves the optimal BRISQUE score and NIQE$\_$S on the SiSRB dataset, significantly outperforming advanced GAN-based architectures. This suggests that while maintaining highly competitive SRI, our framework generates results with fewer artifacts and higher perceptual naturalness.

While RS-GSSR attains competitive CD values, its heavy reliance on large convolutional kernels for boundary smoothing often results in the loss of fine texture details in homogeneous regions. In contrast, our method ensures seamless transitions at shadow boundaries while effectively preserving sharp structural textures. Another distinctive advantage of N$^2$SGSR lies in its processing speed, it attains the highest efficiency among all compared SOTA methods, with an average inference time of approximately $1.3$s for a $512 \times 512$ image. This represents a substantial acceleration compared to complex generative models such as Self-ShadowGAN and SRMF. In conclusion, our N$^2$SGSR not only delivers high-quality shadow removal but also achieves peak processing efficiency among current SOTA benchmarks.

\begin{figure*}[t]
	\centering 
	\includegraphics[width=1\textwidth]{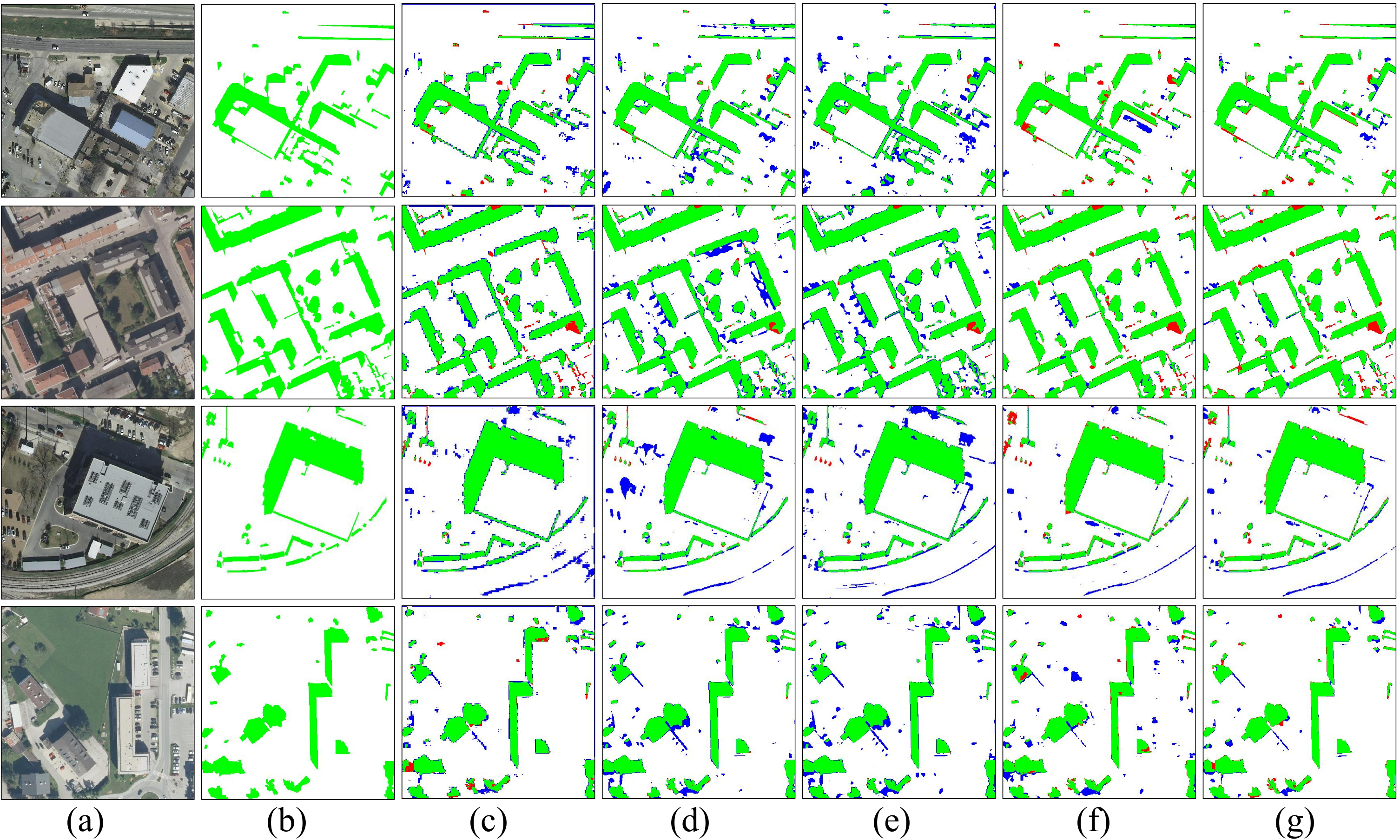}
	\caption{Shadow detection results on the AISD dataset. (a) Input image, (b) Ground truth, (c) RSiSD, (d) AFFPN, (e) ECA-SD, (f) SDCM, (g) RSD, (h) SDDNet, (i) SILT, (j) CADDN, (k) Ours. The rendered colors demote \textcolor{green}{TP(green)}, \textcolor{red}{FN(red)}, \textcolor{blue}{FP(blue)}, TN(white).}
	\label{fig7}
\end{figure*}

\begin{figure*}[t]
	\centering
	\includegraphics[width=1\textwidth]{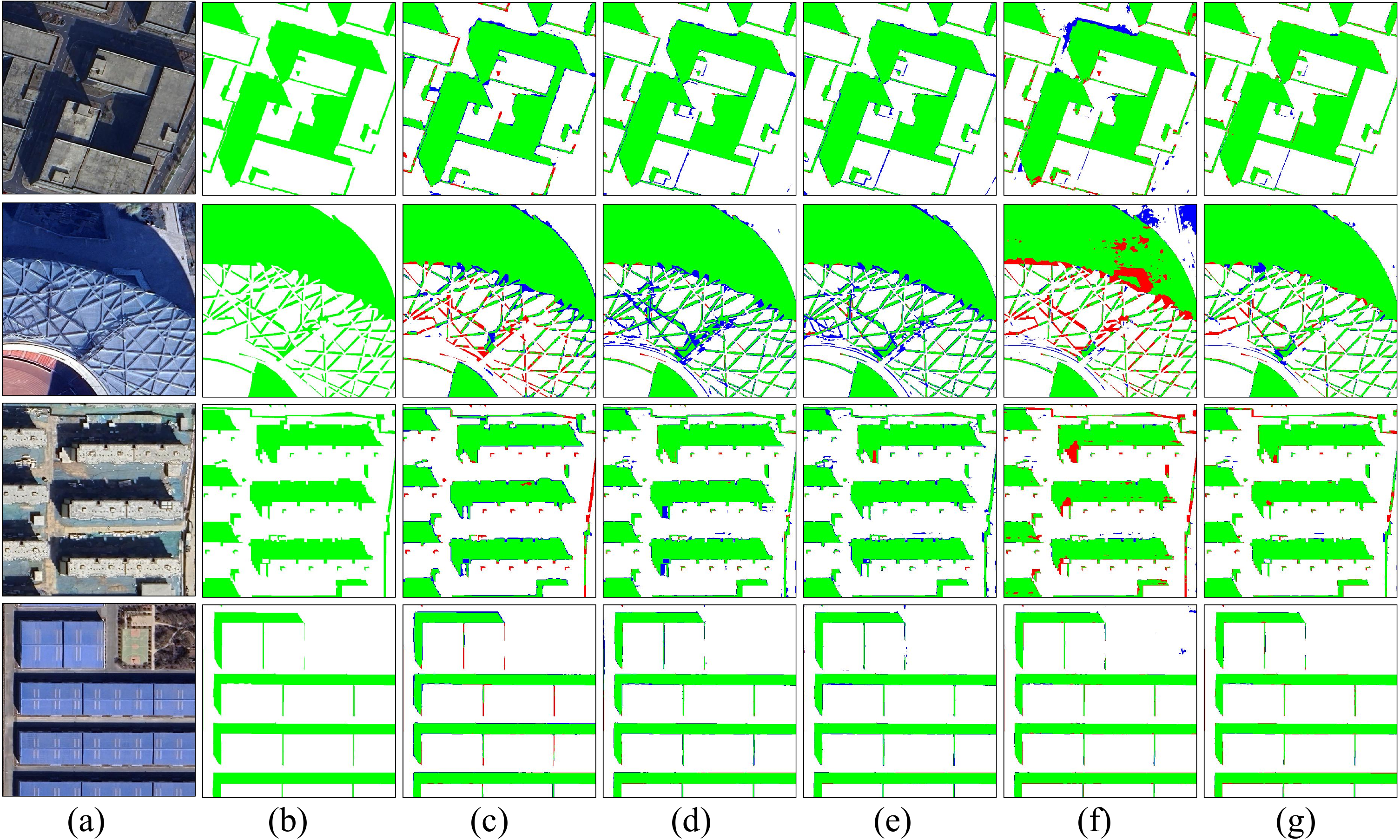}
	\caption{Shadow detection results on our proposed RSISD dataset. (a) Input image, (b) Ground truth, (c) RSD, (d) SDDNet, (e) SILT, (f) CADDN, (g) Ours. The rendered colors demote \textcolor{green}{TP(green)}, \textcolor{red}{FN(red)}, \textcolor{blue}{FP(blue)}, and TN(white).}
	\label{fig8}
\end{figure*}

\subsection{Visualization}
Fig. \ref{fig7} showcases representative scenes from the AISD dataset, characterized by numerous fine-grained shadow boundaries and dark objects that frequently pose challenges for accurate shadow detection. As illustrated in the first row, our method demonstrates precise detection of building-cast shadows while effectively avoiding the common pitfall of misclassifying dark vehicles as shadow regions. The results in the second and fourth rows further highlight our approach's ability to mitigate the shadow discontinuity problem, which often manifests along building edges, thereby ensuring the preservation of integrity and seamless continuity in detected shadow regions. The third row compellingly demonstrates the robust generalization capability of our method: even with imperfect manual annotations, our model consistently identifies subtle slope shadows, showcasing its resilience to minor data inconsistencies.

Similarly, Fig. \ref{fig8} depicts challenging scenarios from the RSISD dataset, known for its extensive building-cast and intricate rooftop shadows. The first row demonstrates our method's success in differentiating between dark road surfaces and genuine shadow regions under low illumination, while accurately detecting shadows cast by gray buildings. For highly complex and densely distributed rooftop shadows, as shown in the second and fourth rows, our method precisely captures fine details while rigorously adhering to physical consistency. Furthermore, the third row provides evidence that in dense residential areas, our approach effectively prevents the erroneous omission of bright objects partially enveloped by shadows, exhibiting exceptional robustness in varied urban environments.

\begin{figure*}[t]
	\centering
	\includegraphics[width=1\textwidth]{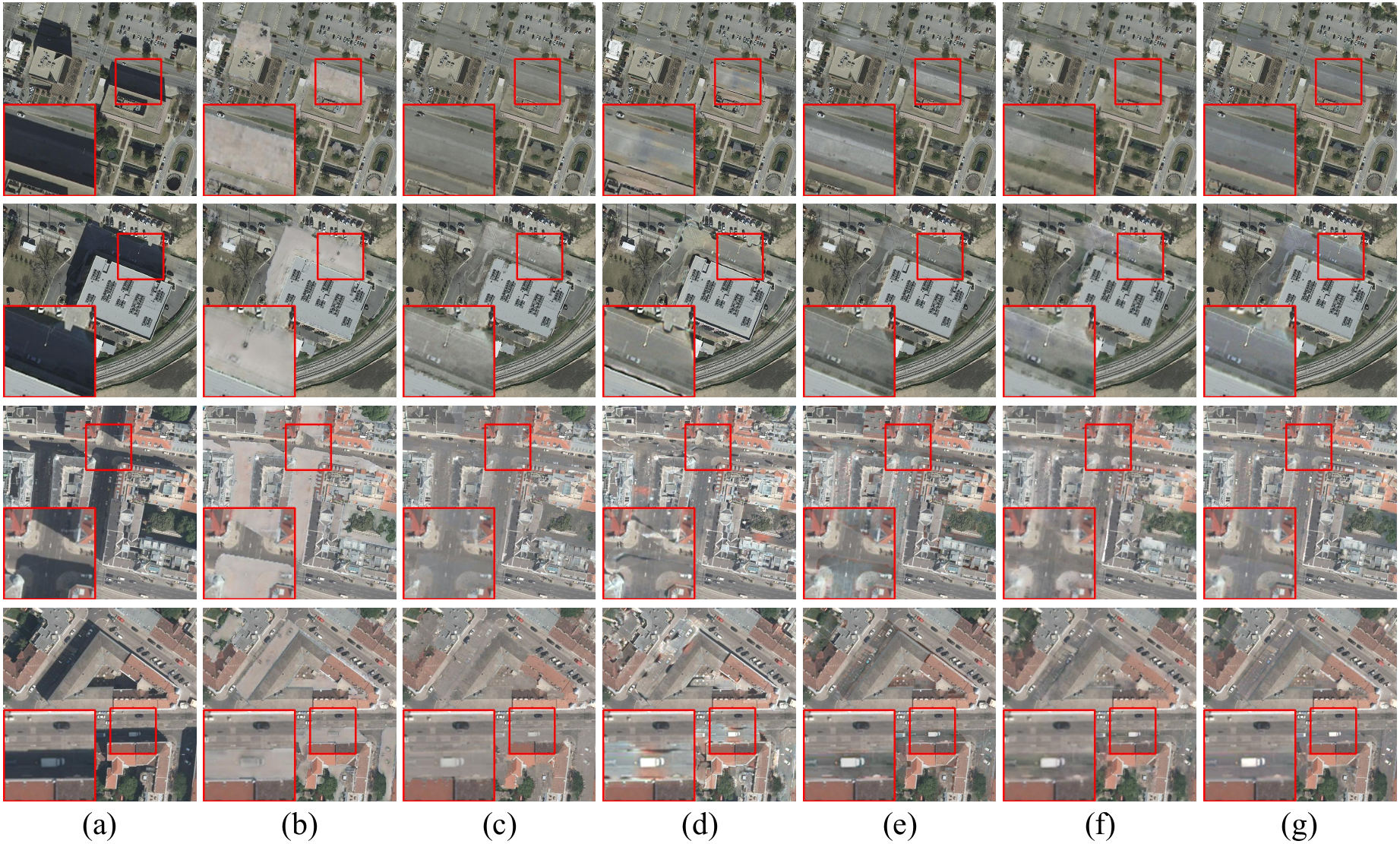}
	\caption{Shadow removal results on the AISD dataset. (a) Input image, (b) G2R-ShadowNet, (c) SRMF, (d) Self-ShadowGAN, (e) MAOSD, (f) RS-GSSR, (g) Ours.}
	\label{fig9}
\end{figure*}

\begin{figure*}[t]
	\centering
	\includegraphics[width=1\textwidth]{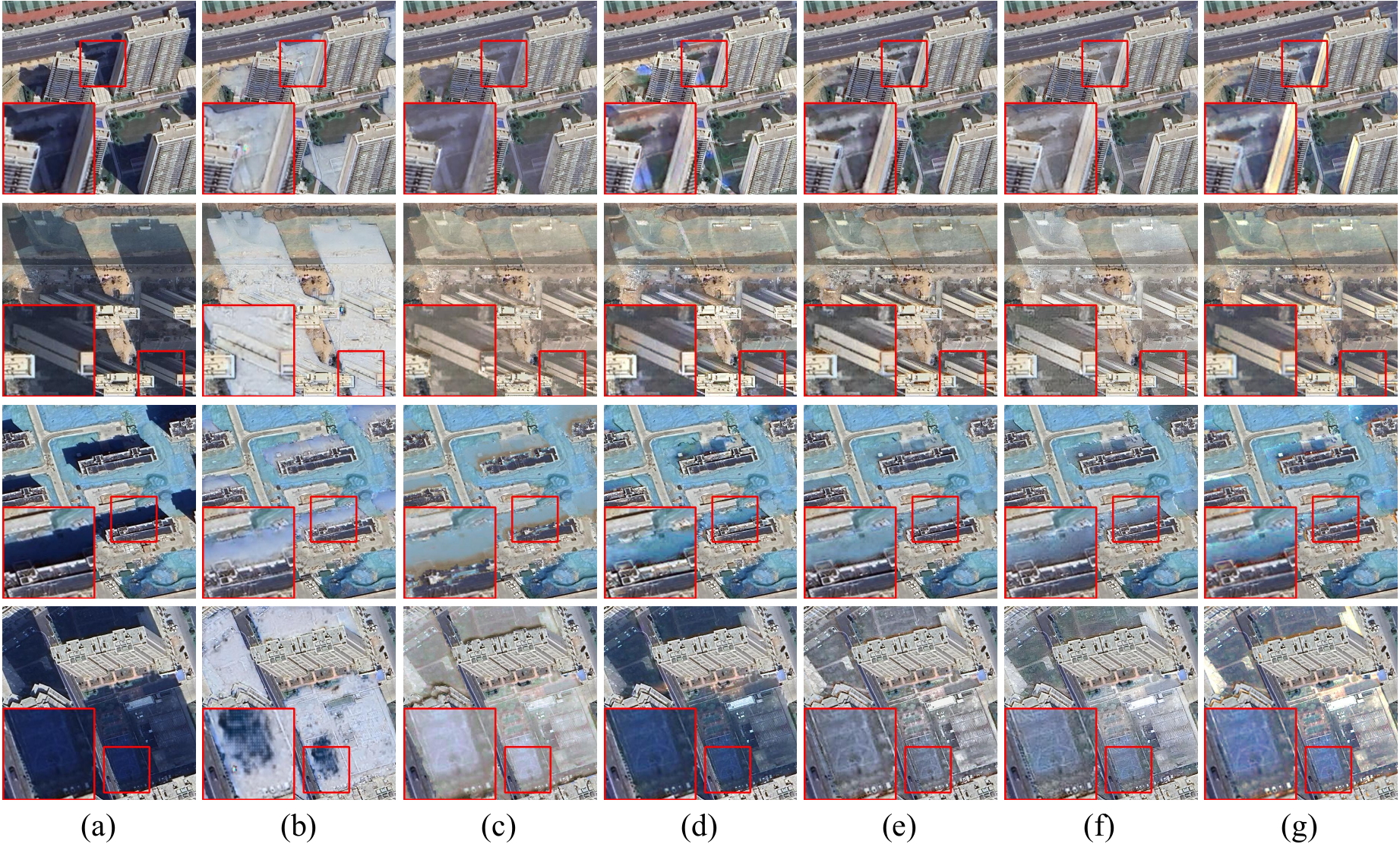}
	\caption{Shadow removal results on our proposed SiSRB dataset. (a) Input image, (b) G2R-ShadowNet, (c) SRMF, (d) Self-ShadowGAN, (e) MAOSD, (f) RS-GSSR, (g) Ours.}
	\label{fig10}
\end{figure*}

Figs. \ref{fig9} and \ref{fig10} provide a visual comparison of shadow removal results on selected scenes from the AISD and SiSRB datasets. G2R-ShadowNet, relying on an adversarial network trained by cropping images, often suffered from a loss of crucial structural information, leading to blurred removal results. The RS-GSSR method, which constructs a paired training dataset through image cropping and shadow synthesis, struggled to robustly model diverse and complex RSI scenarios, occasionally leading to overexposure, as seen in Fig. \ref{fig10}f. In stark contrast, SRMF, Self-ShadowGAN, MAOSD, and our proposed method all ingeniously leverage information from shadow-free regions to guide shadow removal, performing modeling directly on a single input image. Empirical visualizations compellingly indicate that our approach exhibits superior robustness, effectively mitigating common artifacts such as overexposure and color distortion. Moreover, the meticulously designed boundary smoothing mechanism within our method ensures more natural and seamless transitions in penumbra regions, significantly enhancing the visual quality of the images.

\begin{figure*}[t]
	\centering
	\includegraphics[width=1\textwidth]{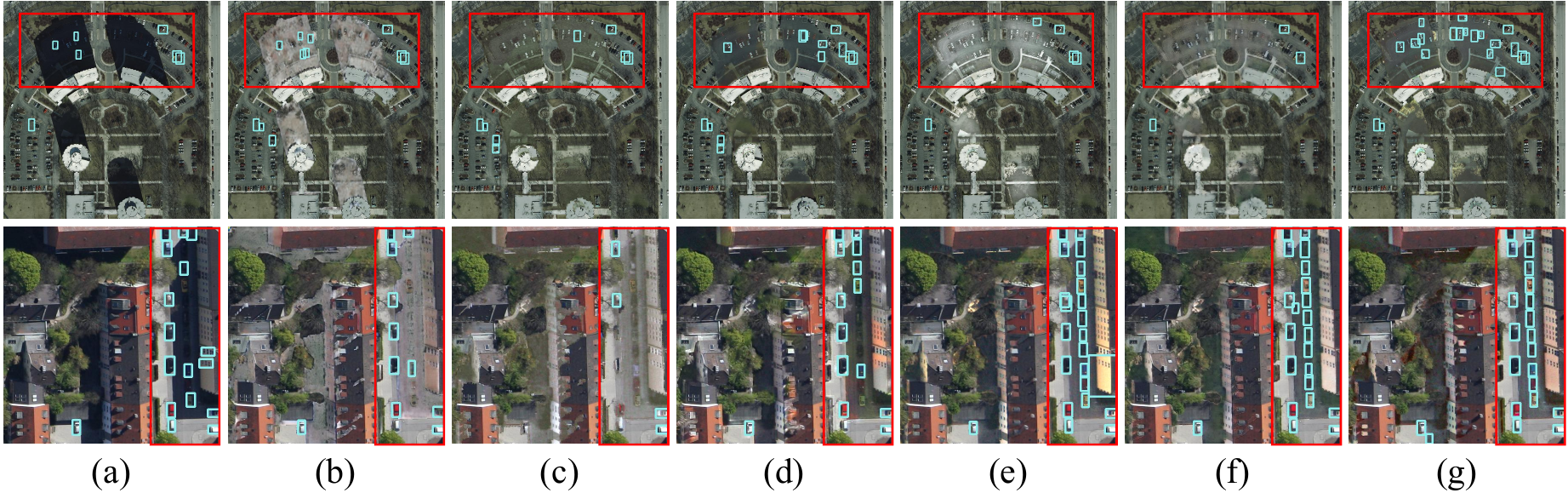}
	\caption{Vehicle detection visualization results for different shadow removal methods. (a) Input image, (b) G2R-ShadowNet, (c) SRMF, (d) Self-ShadowGAN, (e) MAOSD, (f) RS-GSSR, (g) Ours.}
	\label{fig11}
\end{figure*}

To further evaluate the efficacy of shadow removal in enhancing downstream experimental tasks, we employed SCAF-Net~\citep{9530866}, an advanced detection model for aerial imagery, to conduct vehicle detection experiments on the deshadowed images. The quality and accuracy of vehicle detection results serve as an effective indirect metric for assessing the performance of various shadow removal algorithms. Fig.~\ref{fig11} visually presents the vehicle detection outcomes on images processed by different shadow removal methods, with the first row sourced from the AISD dataset and the second row cropped from the DLR3K dataset~\citep{7122912}. Our experimental findings indicate that GAN-based methods, such as G2R-ShadowNet, Self-ShadowGAN, and RS-GSSR, while effective in mitigating shadows, tend to introduce a certain degree of image detail blurring. This blurring artifact subsequently leads to a discernible degradation in vehicle detection performance. This phenomenon is particularly pronounced when detecting small objects like vehicles, where essential textural and edge information crucial for accurate identification might be obscured by the generated artifacts. In contrast, our proposed N$^2$SGSR method demonstrates a significant advantage. N$^2$SGSR effectively and comprehensively leverages the rich information available in non-shadow regions, enabling high-quality shadow removal while meticulously preserving the intrinsic details and structural integrity of vehicle targets within the imagery. As highlighted by the prominent rectangular boxes in Fig.~\ref{fig11}, our method consistently maintains superior clarity and accuracy in vehicle detection. This capability to distinctly capture vehicle contours and positions underscores N$^2$SGSR's exceptional performance in both detail preservation and detection performance enhancement. This advancement holds profound implications for high-precision RSI analysis, highlighting the substantial practical potential of our method.

\subsection{Ablation Study}
\label{subsec:ablation_study}
\subsubsection{Ablation on Shadow Dection}
To thoroughly investigate and quantify the specific contribution of each constituent component within our proposed shadow detection model, a series of granular ablation experiments were systematically conducted on both the AISD and RSISD datasets. The primary focus of these experiments was to evaluate the individual and cumulative impact of the multi-color space features (RGB, HSV, and Lab), the DecoupleNet semantic branch, and the FFM on the overall IoU performance. The results are summarized in Table~\ref{tab:ablation}.
We first analyzed the internal composition of the MCSC module. Using the basic RGB representation as a baseline, the IoU scores were $81.69\%$ on AISD and $71.62\%$ on RSISD. The independent addition of HSV or Lab color spaces led to consistent improvements in detection accuracy on the AISD dataset, confirming that derivative spectral channels provide critical cues for shadow discrimination. Specifically, HSV facilitates the decoupling of luminance from chromaticity, while Lab provides perceptual uniformity that enhances shadow contrast. By integrating all three color spaces, the model achieved a superior IoU of $84.29\%$ on AISD and $72.91\%$ on RSISD, validating that MCSC provides a more comprehensive spectral information for shadows in complex urban scenes.
The impact of the dual-branch architecture was evaluated by comparing the color-only and semantic-only configurations. While the DecoupleNet alone achieved reasonable results, its integration with the MCSC module resulted in a significant performance leap, particularly on the RSISD where the IoU increased from $79.05\%$ to $84.51\%$. This substantial improvement underscores the necessity of combining low-level chromatic variations with high-level semantic context to effectively suppress false positives caused by spectrally similar dark objects.
Finally, the contribution of the FFM was assessed. By comparing the simple concatenation strategy with the full model incorporating FFM, we observed a further increase in IoU to $86.35\%$ on AISD and an impressive $89.71\%$ on RSISD. These findings unequivocally demonstrate the FFM's vital importance in adaptively fusing spatial and semantic features through its cross-spatial attention mechanism, thereby optimizing the final feature representation for precise shadow delineation. In summary, the full DBCSF-Net, incorporating all proposed components, achieves the state-of-the-art performance across both benchmarks.
\begin{table}[h]
	\centering \scriptsize
	\caption{Detailed ablation study of the DBCSF-Net on AISD and RSISD datasets. The symbols $\checkmark$ indicate the inclusion of specific components.}
	\renewcommand{\arraystretch}{1.3}
	\setlength{\tabcolsep}{4pt} 
	\begin{tabular}{ccccccc}
		\hline
		\multicolumn{3}{c}{$\mathrm{MCSC}$} & \multirow{2}{*}{$\mathrm{DecoupleNet}$} & \multirow{2}{*}{$\mathrm{FFM}$} & \multirow{2}{*}{$\mathrm{AISD}$ ($\mathrm{IoU}\uparrow$)} & \multirow{2}{*}{$\mathrm{RSISD}$ ($\mathrm{IoU}\uparrow$)} \\ 
		\cline{1-3}
		$\mathrm{RGB}$ & $\mathrm{HSV}$ & $\mathrm{Lab}$ & & & & \\ 
		\hline
		& & & $\checkmark$ & & 83.78\% & 79.05\% \\
		$\checkmark$ & & & & & 81.69\% & 71.62\% \\
		$\checkmark$ & $\checkmark$ & & & & 83.68\% & 70.91\% \\
		$\checkmark$ & & $\checkmark$ & & & 83.30\% & 70.82\% \\
		$\checkmark$ & $\checkmark$ & $\checkmark$ & & & 84.29\% & 72.91\% \\
		$\checkmark$ & $\checkmark$ & $\checkmark$ & $\checkmark$ & & 84.75\% & 78.46\% \\
		$\checkmark$ & $\checkmark$ & $\checkmark$ & $\checkmark$ & $\checkmark$ & \textbf{86.35\%} & \textbf{89.71\%} \\
		\hline
	\end{tabular}
	\label{tab:ablation}
\end{table}
\subsubsection{Ablation on Shadow Removal}
\begin{figure}
	\centering
	\includegraphics[width=\linewidth]{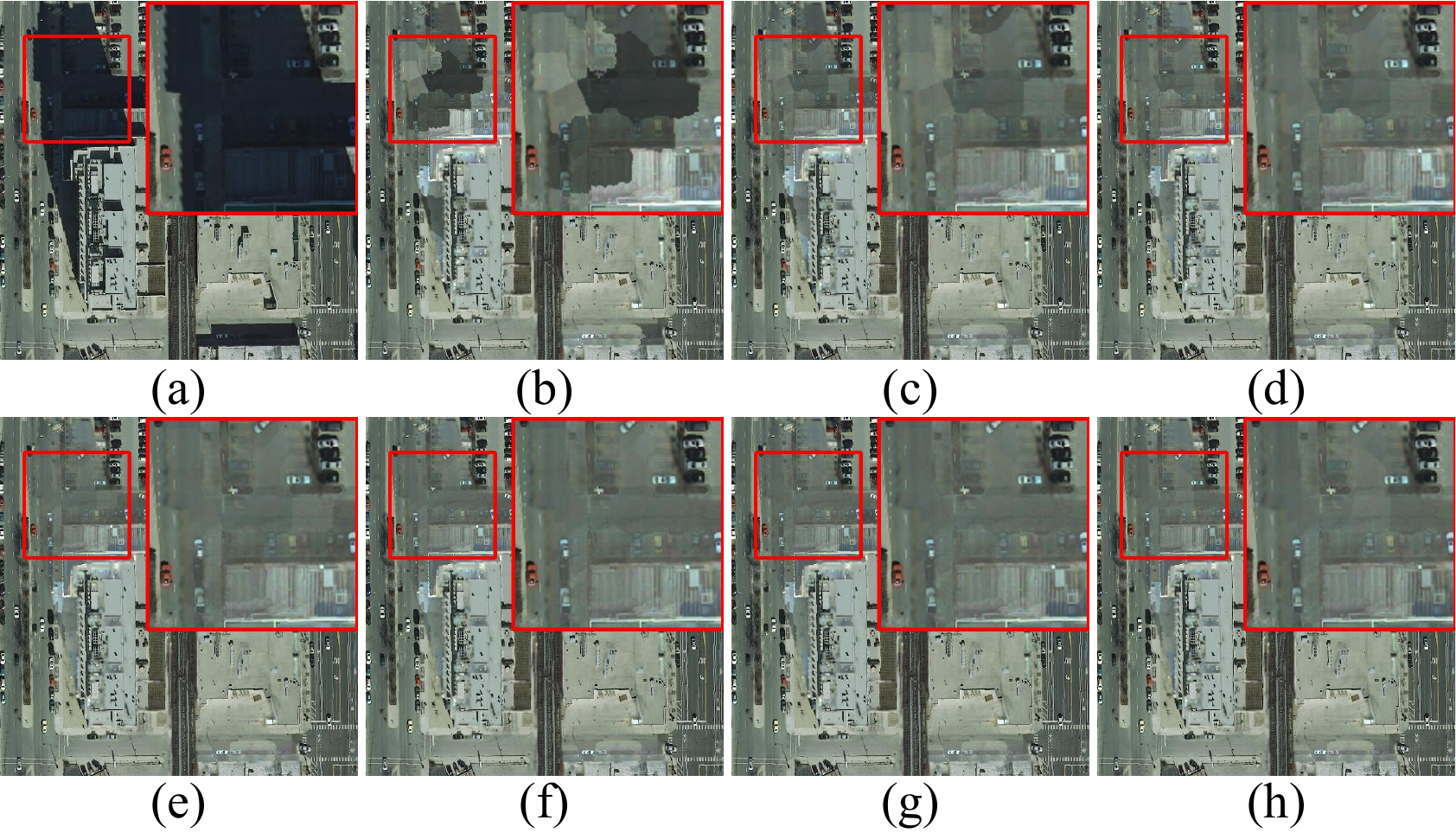}
	\caption{Ablation study on the impact of the number of nearest neighbors $n$ in the $\mathrm{N}^2\mathrm{SGSR}$ algorithm. (a) Input image, (b)--(h) Shadow removal results obtained using $n=1$, $n=3$, $n=5$, $n=7$, $n=9$, $n=12$, and $n=15$, respectively.}
	\label{fign_Ablation}
	\vspace{-2mm}
\end{figure}
The hyperparameter $n$, representing the number of nearest non-shadowed superpixels used for reference, is a critical factor influencing the stability of illumination ratio estimation. As illustrated in Fig. \ref{fign_Ablation}, we conducted a systematic ablation study on the SiSRB dataset by varying $n$, within the range $\{1, 3, 5, 7, 9, 12, 15\}$. When $n=1$, the restoration process relies on a single reference unit, making it highly susceptible to superpixel segmentation noise and local spectral variations, which often results in unstable color recovery. As $n$ increases to seven, the weighted averaging mechanism defined in Equation(\ref{eq19}) effectively suppresses localized noise and leverages neighborhood redundancy, thereby achieving a more natural and consistent spectral transition.
Experimental results indicate that the algorithm performance reaches its optimal peak at approximately $n=7$. Beyond this threshold, the marginal improvement in shadow removal quality becomes negligible, whereas the computational overhead for neighbor searching and weight calculation increases substantially. Furthermore, an excessively large $n$ carries the risk of incorporating distant and semantically irrelevant superpixels, which may slightly degrade the radiometric fidelity. Therefore, to strike an optimal balance between restoration fidelity and computational efficiency, we fixed $n=7$ as the default parameter.
\begin{figure}
	\centering
	\includegraphics[width=\linewidth]{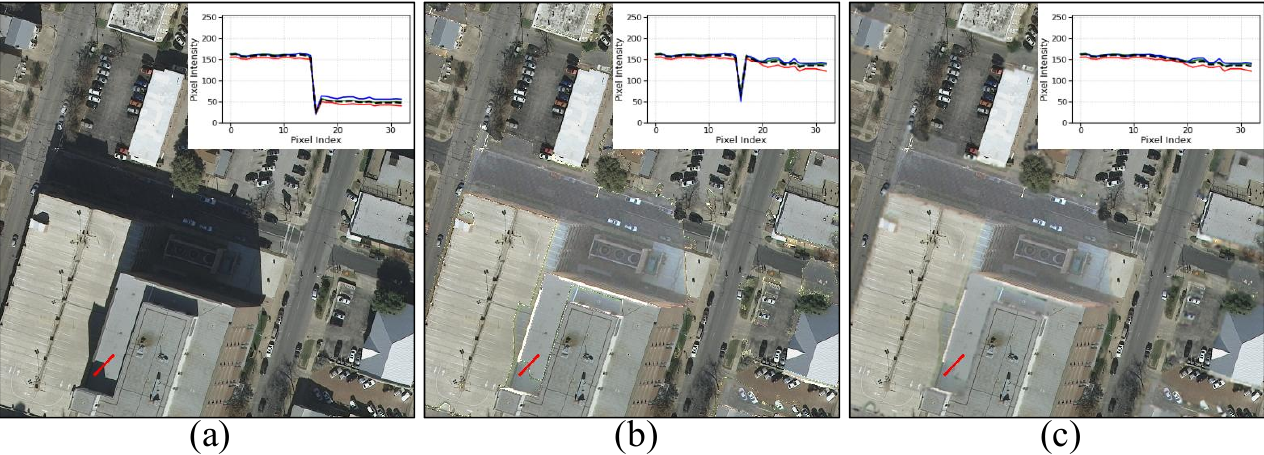}
	\caption{Ablation study on the Bilateral Boundary Smoothing. (a) Input image, (b) Shadow removal result without boundary smoothing, (c) Result with the proposed smoothing mechanism.}
	\label{figBS_Ablation}
	\vspace{-2mm}
\end{figure}
Finally, the contribution of bilateral boundary smoothing to the radiometric fidelity of shadow edges is assessed. Owing to the discrete characteristics of initial binary masks, the direct restoration using estimated illumination ratios frequently induces edge-cliff artifacts. As shown in Fig. \ref{figBS_Ablation}(b), deshadowed regions exhibit stark and unrealistic transitions in the absence of this step. By introducing the bilateral refinement mechanism, the proposed framework facilitates a progressive and physically consistent transition throughout the penumbra. This process effectively mitigates boundary discontinuities and ensures the radiometric integrity of the restored imagery.
\subsection{Analysis of Challenging Scenarios}
\begin{figure}
	\centering
	\includegraphics[width=\linewidth]{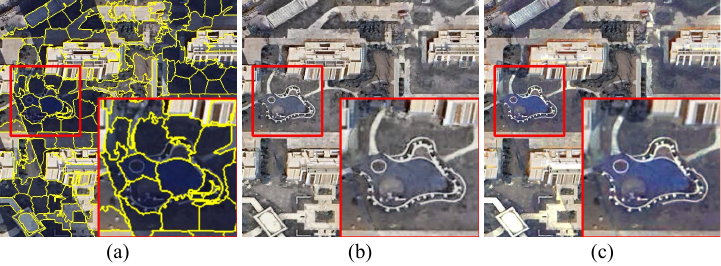}
	\caption{Comparison of different search strategies in extreme scenarios lacking local homogeneous references. (a) Shadowed image after superpixel segmentation. (b) Naive global averaging method. (c) Proposed similarity-weighted global search method.}
	\label{fig13}
	\vspace{-2mm}
\end{figure}
Finally, we investigate the performance of N$^2$SGSR in extreme scenarios characterized by a deficiency of local homogeneous references. Fig. \ref{fig13} illustrates a challenging case where an extensive shadow completely occludes a pool, and no similar land-cover types exist within the adjacent shadow-free neighborhood.
In this experiment, we compare two distinct processing strategies for the shadowed pool region: naive global averaging and similarity-weighted global search. As shown in the zoomed-in regions of Fig. \ref{fig13}(b), the naive method produces conspicuous chromatic artifacts, as it indiscriminately incorporates radiometric information from heterogeneous materials (e.g., surrounding rooftops). In contrast, as shown in Fig. \ref{fig13}(c), the similarity-weighted global search effectively retrieves spectrally similar candidates from the entire scene to restore the realistic appearance of the pool. Although the absence of direct category-specific spectral samples renders this reconstruction inherently an ill-posed problem, our method maintains a superior degree of radiometric consistency by leveraging the global radiometric context of the scene rather than relying on biased heterogeneous neighbors. We anticipate that integrating high-dimensional semantic retrieval could further refine these results, which remains a promising direction for future research.

\section{Conclusion and Future Work}
This paper presented SARU, which unifies the tasks of shadow detection and removal, addressing the critical challenges of cumulative error propagation and the reliance on paired training data. Our main contribution is a synergistic pipeline that integrates a high-fidelity detection network (DBCSF-Net) to prevent misclassification of dark objects, with a novel, training-free removal algorithm (N$^2$SGSR) that eliminates the need for paired data. To support this, we also introduce two newly constructed benchmarks, RSISD and SiSRB, to foster future research.
Extensive experiments demonstrate that our approach achieves state-of-the-art performance. By decoupling high-quality restoration from the need for paired ground truth, SARU offers a practical and scalable solution for enhancing RSI in real-world applications. Future work will focus on improving adaptability to dynamic illumination and extending our physically-based principles to other vision tasks, unlocking broader capabilities for a wide range of applications.

\section*{Acknowledgements}
This work was supported in part by the NSFC Key Project of Joint Fund for Enterprise Innovation and Development under Grant U24A20342, and in part by the National Natural Science Foundation of China under Grant 62576006 and 61976004.

\printcredits

\bibliography{SARU}

\end{sloppypar}
\end{document}